\documentclass[journal]{IEEEtran}
\ifCLASSINFOpdf
\else
\fi
\usepackage{graphicx}
\usepackage{multicol}
\usepackage{multirow}
\usepackage[bookmarks=true]{hyperref}
\usepackage{amssymb,amsmath}
\usepackage{subfig}
\usepackage{bbding}

\newcommand{\eq}{Eq.~}

\newcommand{\rl}{L}
\newcommand{\om}{O}
\newcommand{\cm}{M}

\newcommand{\hc}{C}
\newcommand{\qd}{Q}

\newcommand{\pdf}{\mathbf{pdf}}

\newcommand{\argmax}[1]{\underset{#1}{\operatorname{argmax}}\medspace}

\begin{document}

%
\title{Deep Dexterous Grasping of Novel Objects from a Single View}
%
%
%

\author{Umit Rusen Aktas$^{1}$, Chao Zhao$^{1}$, Marek Kopicki$^{1}$, Ales Leonardis$^{1}$ and Jeremy L. Wyatt$^{1}$
\thanks{*This work was primarily supported by FP7-ICT-600918}
\thanks{$^{1}$ University of Birmingham, School of Computer Science, UK.
        {\tt\small jeremy.l.wyatt@gmail.com}}%
}

%
%

\markboth{IEEE Transactions on Robotics}%
{Shell \MakeLowercase{\textit{et al.}}: Deep Dexterous Grasping}
%



\maketitle

\begin{abstract}
Dexterous grasping of a novel object given a single view is an open problem. This paper makes several contributions to its solution. First, we present a simulator for generating and testing dexterous grasps. Second we present a data set, generated by  this simulator, of 2.4 million simulated dexterous grasps of variations of 294 base objects drawn from 20 categories. Third, we present a basic architecture for generation and evaluation of dexterous grasps that may be trained in a supervised manner. Fourth, we present three different evaluative architectures, employing ResNet-50 or VGG16 as their visual backbone. Fifth, we train, and evaluate seventeen variants of generative-evaluative architectures on this simulated data set, showing improvement from 69.53\% grasp success rate to 90.49\%. Finally, we present a real robot implementation and evaluate the four most promising variants, executing 196 real robot grasps in total. We show that our best architectural variant achieves a grasp success rate of 87.8\% on real novel objects seen from a single view, improving on a baseline of 57.1\%. 

\end{abstract}

\begin{IEEEkeywords}
Deep learning, generative-evaluative learning, grasping.
\end{IEEEkeywords}

%
\IEEEpeerreviewmaketitle

\section{Introduction}
%
%

If robots are to be widely deployed in human populated environments then they must deal with unfamiliar situations. An example is the case of grasping and manipulation. Humans grasp and manipulate hundreds of objects each day, many of which are previously unseen. Yet humans are able to dexterously grasp these novel objects with a rich variety of grasps. In addition, we do so from only a single, brief, view of each object. To operate in our world, dexterous robots must replicate this ability.

This is the motivation for the problem tackled in this paper, which is planning of (i) a dexterous grasp, (ii) for a novel object, (iii) given a single view of that object. We define dexterous as meaning that the robot employs a variety of dexterous grasp types across a set of objects. The combination of constraints (i)-(iii) makes grasp planning hard because surface reconstruction will be partial, yet this cannot be compensated for by estimating pose for a known object model. The novelty of the object, together with incomplete surface reconstruction, and uncertainty about object mass and coefficients of friction, renders infeasible the use of grasp planners which employ classical mechanics to predict grasp quality. Instead, we must employ a learning approach.

This in turn raises the question as to how we architect the learner. Grasp planning comprises two problems: generation and evaluation. Candidate grasps must first be generated according to some distribution conditioned on sensed data. Then each candidate grasp must be evaluated, so as to produce a grasp quality measure (e.g maximum resistable wrench), the probability of grasp success, the likely in-hand slip or rotation, etcetera. These measures are then used to rank grasps so as to select one to execute.
\begin{figure}[t]
\begin{center}
  \includegraphics[width=\columnwidth]{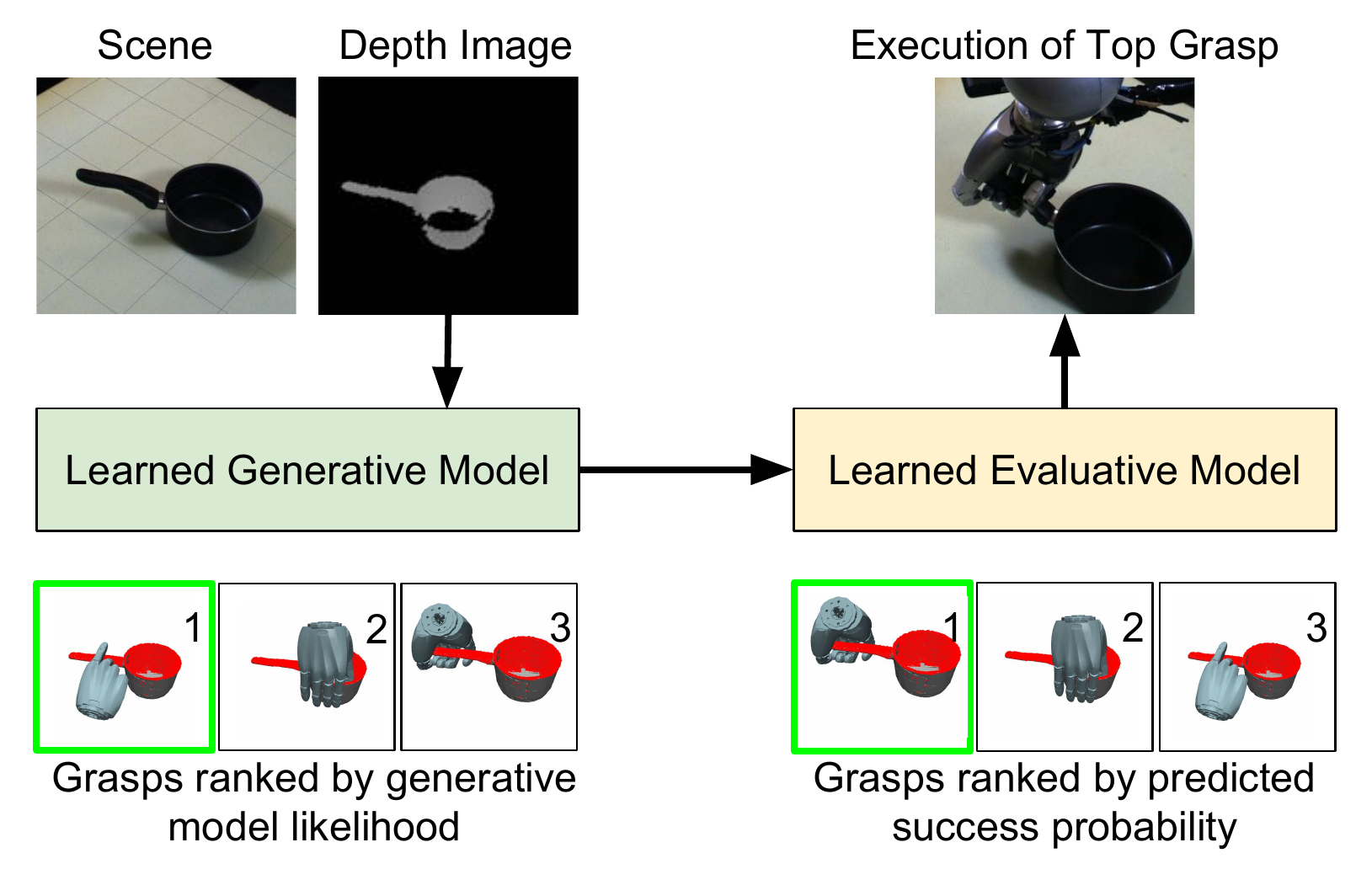}
  \end{center}
  \caption{The basic architecture of a generative-evaluative learner. When shown a novel object the learned generative model (GM) produces many grasps according to its likelihood model. These are then each evaluated by a learned evaluative model (EM), which predicts the probability of grasp success. The grasps are then re-ranked according to the predicted success probability and the top ranked grasp is executed.}
\label{fig:systemArchitecture}
\end{figure}
Either or both a {\em generative} or {\em evaluative} model may be learned. If only a generative model is learned then evaluation must be carried out using mechanically informed reasoning, which, as we noted, cannot easily be applied to the case of novel objects seen from a single view. If only an evaluative model is learned then grasp generation must proceed by search. This is challenging for true dexterous grasping as the hand may have between nine and twenty actuated degrees of freedom. Thus, for dexterous grasping of novel objects from a single view, it becomes appealing to {\em learn} both the generative and the evaluative model. 

The contributions of this paper are as follows. First, we present a data-set of 2.4 million dexterous grasps in simulation that may be used to evaluate dexterous grasping algorithms. Second, we release the source code of the dexterous grasp simulator, which can be used to visualise the dataset and gather new data.\footnote{The code and simulated grasp dataset are available at \href{https://rusen.github.io/DDG}{https://rusen.github.io/DDG}. The web page explains how to download the dataset, install the physics simulator and re-run the grasps in simulation. The simulator acts as a client alongside a simple web server to gather new grasp data in a distributed setup.} Third, we present a generative-evaluative architecture that combines data efficient learning of the generative model with data intensive learning in simulation of an evaluative model. Fourth, we present multiple variations of the evaluative model. Fifth, we present an extensive evaluation of all these models on our simulated data set. Finally, we compare the two most promising variants on a real robot with a data-set of objects in challenging poses.

The model variants are organised in three dimensions. First, we employ two different generative models (GM1 \cite{kopicki2015ijrr} and GM2 \cite{kopicki2019ijrr}), one of which (GM2) is designed specifically for single view grasping. Second, we use two different back-bones for the evaluative model, VGG-16 and ResNet-50. Third, we experiment with two optimisation techniques--gradient ascent (GA) and stochastic annealing (SA)--to search for better grasps using the evaluative model as an objective function.



The paper is structured as follows. First, we discuss related work. Second, the basic generative model is described in detail and the main features of the extended generative model are sketched. Third, we describe the design of the grasp simulation, the generation of the data set. Fourth, we describe the different architectures employed for the evaluative model. Fifth, we describe the evaluative model training, the optimisation variants for the evaluative model and the simulated experimental study. Finally, we present the real robot study.

\section{Background and Related Work}

There are four broad approaches to grasp planning. First, we may employ analytic mechanics to evaluate grasp quality. Second, we may engineer a mapping from sensing to grasp. Third, we may learn this mapping, such as learning a generative model. Fourth, we may learn a mapping from sensing and a grasp to a grasp success prediction. See \cite{bohg2014data} and  \cite{sahbani2012overview} for recent reviews of data driven and analytic methods respectively.

Analytic approaches use mechanical models to predict grasp outcome \cite{bicchi2000a,Liu2000,Pollard2004,miller2004}. This requires models of both object (mass, mass distribution, shape, and surface friction) and manipulator (kinematics, exertable forces and torques). Several grasp quality metrics can be defined using these~\cite{Ferrari1992,Roa2015,Shimoga1996} under a variety of mechanical assumptions. These have been applied to dexterous grasp planning \cite{Boutselis2014,Gori2014,Hang2014,Rosales2012,Saut2012,ciocarlie2009hand}. The main drawback of analytic approaches is that estimation of object properties is hard. Even a small error in estimated shape, friction or mass will render a grasp unstable \cite{zheng2005a}. There is also evidence that grasp quality metrics are not well correlated with actual grasp success \cite{bekiroglu2011b,kim2013a,goins2014a}.

An alternative is learning for robot grasping, which has made steady progress. There are probabilistic machine learning techniques employed for surface estimation for grasping \cite{dragiev2011gaussian}; data efficient methods for learning dexterous grasps from demonstration \cite{ben-amor2012a,kopicki2015ijrr,Osa2018}; logistic regression for classifying grasp features from images \cite{saxena2008a}; extracting generalisable parts for grasping \cite{detry2012a} and for autonomous grasp learning \cite{detry2010a}. Deep learning is a recent approach to grasping. Most work is for two finger grippers. Approaches either learn an evaluation function for an image-grasp pair \cite{levine16,lenz2015deep,gualtieri2016high,mahler2017dex,pinto2016supersizing,johns2016deep}, learn to predict the grasp parameters \cite{redmon2015real,kumra2017iros} or jointly estimate both \cite{morrison18}. The quantity of real training grasps can be reduced by mixing real and simulated data \cite{bousmalis2017using}.

\begin{table*}[t]
\centering
\begin{tabular}{|l|l|l|l|l|l|l|l|}
\hline
References & \multicolumn{3}{|c|}{Grasp type} & Robot & Clutter & Model & Novel  \\ \cline{2-4}
                   & 2-fing. & $>$2-finger & $>$2-finger & results & & free & objects\\ 
                   &            & power           & dexterous &           &  &        &           \\ \hline
\cite{detry2012a,saxena2008a,detry2010a,lenz2015deep,mahler2017dex,johns2016deep,morrison-RSS-18} & \Checkmark & & & \Checkmark &  & \Checkmark & \Checkmark \\ \hline      
\cite{pinto2016supersizing,bousmalis2017using,levine2017,Gualtieri2016} & \Checkmark & & & \Checkmark & \Checkmark & \Checkmark & \Checkmark \\ \hline     
\cite{kappler2015leveraging} &  & \Checkmark & &  &  &  & \Checkmark \\ \hline
\cite{zhou20176dof} &  & \Checkmark & &  &  & \Checkmark & \Checkmark \\ \hline     
\cite{lu2017planning,varley2015generating} &  & \Checkmark & & \Checkmark &  & \Checkmark & \Checkmark \\ \hline    
\cite{ben-amor2012a} &  & &  \Checkmark &  \Checkmark &  &  & \Checkmark \\ \hline    
\cite{veres2017modeling,zhou20176dof,kappler2015leveraging} &  & &  \Checkmark &   &  & \Checkmark & \Checkmark \\ \hline 
\cite{kopicki2015ijrr,kopicki2019ijrr,Osa2018} &  & &  \Checkmark &  \Checkmark &  & \Checkmark & \Checkmark \\ \hline 
\cite{arruda2016active} &  & &  \Checkmark &  \Checkmark & \Checkmark & \Checkmark & \Checkmark \\ \hline 
This paper &  & &  \Checkmark &  \Checkmark &  & \Checkmark & \Checkmark \\ \hline 
\end{tabular}
\caption{Qualitative comparison of grasp learning methods.}
\label{tab:comp-related-work}
\end{table*}

A small number of papers have explored deep learning as a method for dexterous grasping. \cite{lu2017planning,varley2015generating,veres2017modeling,zhou20176dof,kappler2015leveraging}. All of these use simulation to generate the training set for learning. Kappler \cite{kappler2015leveraging} showed the ability of a CNN to predict grasp quality for multi-fingered grasps, but uses complete point clouds as object models and only varies the wrist pose for the pre-grasp position, leaving the finger configurations the same. Varley \cite{varley2015generating} and later Zhou \cite{zhou20176dof} went beyond this by varying the hand pre-shape, and predicting from a single image of the scene. Each of these posed search for the grasp as a pure optimisation problem (using simulated annealing or quasi-Newton methods) on the output of the CNN. They, also, take the approach of learning an evaluative model, and generate candidates for evaluation uninfluenced by prior knowledge. Veres \cite{veres2017modeling}, in contrast, learns a deep generative model. Finally Lu \cite{lu2017planning} learns an evaluative model, and then, given an input image, optimises the inputs that describe the wrist pose and hand pre-shape to this model via gradient ascent, but does not learn a generative model. In addition, the grasps start with a heuristic grasp which is varied within a limited envelope. Of the papers on dexterous grasp learning with deep networks only two approaches \cite{varley2015generating,lu2017planning} have been tested on real grasps, with eight and five test objects each, producing success rates of 75\% and 84\% respectively. An key restriction of both of these methods is that they only plan the pre-grasp, not the finger-surface contacts, and are thus limited to power-grasps.

Thus, in each case, either an evaluative model is learned but there is no learned prior over the grasp configuration able to be employed as a generative model; or a generative grasp model is learned, but there is no evaluative model learned to select the grasp. Our technical novelty is thus to bring together a data-efficient method of learning a good generative model with an evaluative model. As with others, we learn the evaluative model from simulation, but the generative model is learned from a small number of demonstrated grasps. Table~\ref{tab:comp-related-work} compares the properties of the learning methods reviewed above against this paper. Most works concern pinch grasping. Of the eight papers on learning methods for dexterous grasping, two \cite{varley2015generating, lu2017planning} are limited to power grasps. Of the remaining five, three have no real robot results \cite{veres2017modeling,zhou20176dof,kappler2015leveraging}. Of the remaining four, two we directly build on here, the third being a extension of one of those grasp methods with active vision. Finally, our real robot evaluation is extensive in comparison with competitor works on dexterous grasping, comprising 196 real grasps of 40 different objects.

\section{Data Efficient Learning of a Generative Grasp Model from Demonstration}

This section describes the generative model learning upon which the paper builds. We employ two related grasp generation techniques \cite{kopicki2015ijrr, kopicki2019ijrr}, which both learn a generative model of a dexterous grasp from a demonstration (LfD). Those papers both posed the problem as one of learning a factored probabilistic model from a single example. The method is split into a model learning phase, a model transfer phase, and the grasp generation phase. 

\subsection{Model learning}
The model learning is split into three parts: acquiring an {\em object model}; using this object model, with a demonstrated grasp, to build a {\em contact model} for each finger link in contact with the object; and acquiring a {\em hand configuration model} from the demonstrated grasp. After learning the object model can be discarded.

\subsubsection{Object model}
First, a point cloud of the object used for the demonstrated grasp is acquired by a depth camera, from several views. Each point is augmented with the estimated principal curvatures at that point and a surface normal. Thus, the $j^{th}$ point  in the cloud gives rise to a feature $x_j=(p_j, q_j, r_j)$, with the components being its position $p_j \in \mathbb R^3$, orientation $q_j \in SO(3)$ and principal curvatures $r_j=(r_{j,1},r_{j,2}) \in \mathbb R^2$. The orientation $q_j$ is defined by $k_{j,1},k_{j,2}$, which are the directions of the principal curvatures.  For later convenience we use $v=(p,q)$ to denote position and orientation combined. These features $x_j$ allow the object model to be defined as a kernel density estimate of the joint density over $v$ and $r$.
\begin{equation}
\om(v, r) \equiv \pdf^\om(v, r) \simeq \sum_{j=1}^{K_O} w_j \mathcal{K}(v, r|{x_j}, \sigma_{x})
\label{eq:om}
\end{equation}
where $\om$ is short for $\pdf^\om$, bandwidth $\sigma_{x} = (\sigma_{p}, \sigma _{q}, \sigma_{r})$, $K_O$ is the number of features $x_j$ in the object model, all weights are equal $w_j = 1/{K_O}$, and $\mathcal{K}$ is defined as a product:
\begin{equation}\label{eq:kernel_in_se3}
\mathcal{K}(x | \mu, \sigma) = \mathcal{N}_3(p| \mu_p, \sigma_p) \Theta(q| \mu_q, \sigma_q) \mathcal{N}_2(r| \mu_r, \sigma_r)
\end{equation}
where $\mu$ is the kernel mean point, $\sigma$ is the kernel bandwidth, $\mathcal{N}_n$ is an $n$-variate isotropic Gaussian kernel, and ${\Theta}$ corresponds to a pair of antipodal von Mises-Fisher distributions.
\begin{figure*}[t]
\includegraphics[width=\textwidth]{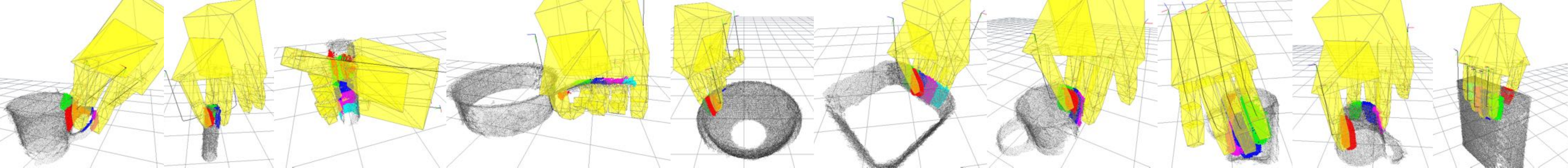}
\caption{The ten training grasps for the generative model. The final hand pose is shown in yellow, the sensed point cloud in black, and the parts of the point cloud that contribute to each contact model are coloured by the associated link. \label{fig:generative-training}}
\end{figure*}
\subsubsection{Contact models}
When a grasp is demonstrated the final hand pose is recorded. This is used to find all the finger links $L$ and surface features $x_j$ that are in close proximity. A contact model $M_i$ is built for each finger link $i$. Each feature in the object model that is within some distance $\delta_i$ of finger link $L_i$ contributes to the contact model $\cm_i$ for that link. This contact model is defined for finger link $i$ as follows:
\begin{equation}
\cm_i(u, r) \equiv \pdf^\cm_i(u, r) \simeq \frac{1}{Z} \sum_{j=1}^{K_{M_i}} w_{ij} \mathcal{K}(u, r | {x_j}, \sigma_{x})
\label{eq:cm}
\end{equation}
where $u$ is the pose of $\rl_i$ relative to the pose $v_j$ of the $j^{\mathnormal{th}}$ surface feature, $K_{M_i}$ is the number of surface features in the neighbourhood of link $L_i$, $Z$ is the normalising constant, and $w_{ij}$ is a weight that falls off exponentially as the distance between the feature $x_j$ and the closest point $a_{ij}$ on finger link $L_i$ increases:
\begin{equation}
w_{ij} = \begin{cases}\exp(-\lambda ||p_j-a_{ij}||^2) \quad &\textnormal{ if } ||p_j-a_{ij}|| < \delta_i\\
0 \quad &\textnormal{ otherwise},\end{cases}
\label{eq:learning.modeldist.wgh}
\end{equation}
The key property of a contact model is that it is conditioned on local surface features likely to be found on other objects, so that the grasp can be transferred. We use the principal curvatures $r$, but many local surface descriptors would do. 
%

\subsection{Hand configuration model}
In addition to a contact model for each finger-link, a model of the hand configuration $h_c \in \mathbb R^D$ is recorded, where $D$ is the number of DoF in the hand. $h_c$  is recorded for several points on the demonstrated grasp trajectory as the hand closed. The learned model is:
\begin{equation}
\hc(h_c) \equiv \sum_{\gamma \in [-\beta, \beta]} w({h_c(\gamma)}) \mathcal{N}_D(h_c|h_c(\gamma), \sigma_{h_c}) 
\label{eq:hc}
\end{equation}
where $w({h_c(\gamma)}) = \exp(-\alpha \|h_c(\gamma) - h^g_c \|^2)$; $\gamma$ is a parameter that interpolates between the beginning ($h^t_c$) and end ($h^g_c$) points on the trajectory, governed via \eq\ref{eq:learning.configmodel.config} below; and $\beta$ is a parameter that allows extrapolation of the hand configuration.
\begin{equation}
h_c(\gamma) = (1 - \gamma)h^g_c + \gamma h^t_c
\label{eq:learning.configmodel.config}
\end{equation}
\subsection{Grasp Transfer}
When presented with a new object $o_{new}$ the contact models must be transferred to that object. A partial point cloud of $o_{new}$ is acquired (from a single view) and recast as a density, $\om_{new}$, again using \eq \ref{eq:om}. The transfer of each contact model $\cm_i$ is achieved by convolving $\cm_i$ with $\om_{new}$. This convolution is approximated with a Monte-Carlo method, resulting in an kernel density model of the pose $s$ of the finger link $i$ (in workspace coordinates) for the new object. The Monte-Carlo procedure samples poses for link $L_i$ on the new object. The $j^{th}$ sample is $\hat{s}_{ij}=(\hat{p}_{ij},\hat{q}_{ij})$. Each sample $\hat{s}_{ij}$ is weighted $w_{ij}$ by its likelihood. These samples are used to build what we term the query density:
\begin{equation}
\qd_i(s) \simeq \sum^{K_{Q_i}}_{j=1} w_{ij} \mathcal{N}_3(p|{\hat{p}_{ij}}, \sigma_{p}) \Theta(q|{\hat{q}_{ij}}, \sigma_{q})
\label{eq:qd.approx}
\end{equation}
where all the weights are normalised, $\sum_j w_{ij} = 1$. A query density is constructed for every contact model and the new object. These query densities, together with the hand configuration model, are then used to generate grasps. Query density computation is fast, taking $<0.5s$  per grasp model.

\subsection{Grasp generation}
Given a set of query densities and hand configuration models, candidate grasps may be generated as follows. Select a query density $k$ a random and take a sample  for a finger link pose on the new object $s_k \sim \qd_{k}$. Then, take a sample $h_c \sim C$ from the hand configuration model. This pair of samples together define, via the hand kinematics, a complete grasp $h=(h_w,h_c)$, where $h_w$ is the pose of the wrist and $h_c$ is the configuration of the hand. The initial grasp is then improved by stochastic hill-climbing on a product of experts:
\begin{equation}
\argmax{(h_w, h_c)} \hc(h_c) \prod_{\qd_i \in \mathcal{Q}} \qd_i\left(k_{i}^{\mathrm{for}}\left(h_w, h_c\right)\right)
\label{eq:grasping.product}
\end{equation}
This generate and improvement process has periodic pruning steps, in which only the higher likelihood grasps are retained. It can be run many times, thus enabling the generation of many candidate grasps. In addition, a separate generative model can be learned for each demonstrated grasp. Thus, when presented with a new object, each grasp model can be used to generate and improve grasps. We typically generate and optimise 100 grasps per grasp type. Finally, the many candidate grasps generated from each grasp model can be compared and ranked according to their likelihoods. The product of experts formulation, however, only ensures that the generated grasps have high likelihood according to the model. There is no estimate of the probability that the grasp will succeed. This motivates the dual architecture in this paper. This completes the description of our first generative model, which we refer to as GM1. We now proceed to quickly outline the extensions made to GM1 so as to produce GM2. \label{section:generative}

\section{Improved Generative Learning}


In this paper we also utilised a more advanced generative model, which we refer to as GM2. This model has three features which are different from the base model GM1. As for GM1, these are not a contribution of this paper and are described fully in \cite{kopicki2019ijrr}. For completeness, however, we briefly describe the three differences between GM2 and GM1. 

\subsection{Object View Model}\label{sec:representations.object}
The first difference is that the learning of grasp models is done per view, rather than per grasp. For a training grasp made on an object viewed from seven viewpoints, there will be seven grasp models learned. This enables grasps to generalise better when the testing object to be grasped is thick and is only seen from a single view. The view based models allow a greater role to be played by the hand shape model and this enables generated grasps to have fingers which `float' behind a back surface that cannot be seen by the robot.

\subsection{Clustering Contact Models}\label{sec:learning.clustering}

The second innovation is the ability to merge grasp models learned from different grasps. In the memory based scheme of GM1, the number of contact models $N_{\mathcal{M}}$ equals the product of the number of training grasps by the number of views. This has two undesirable properties. First, it means that generation of grasps for test objects rises linearly in the number of training grasps. Second, it limits the generalisation power of the contact models. We can overcome these problems by clustering the contact models from each training grasp. To do this we need a measure of the similarity between any pair of contact models. Recall that our contact models are probability densities represented as kernel density estimators. Thus, we need a distance metric in the space of probability densities of a given dimension.

One possibility is to employ Jensen-Shannon distance, but this is slow to evaluate. We therefore start by devising a simple and quick to compute asymmetric divergence. We then build on top of it a symmetric distance. Having obtained this distance measure we can employ our clustering method of choice, which in our case was affinity propagation \cite{frey2007clustering}. After clustering, we compute a cluster prototype as described in \cite{kopicki2019}.

\subsection{Improved Grasp Transfer and Inference}
GM2 utilises the same distance measure to transfer grasps when creating the query densities and also to evaluate candidate grasps. This has the effect of making the proposed grasps more conservative and thus closer to the demonstrated grasps in terms of the type of contacts made with the target object.

We now proceed to describe how we use these models to generate a data-set of 2 million simulated dexterous grasps. \label{section:generative_new}



\section{The Simulated Grasp Data Set}
\label{section:simulation}
In this section, we describe how we generated a realistic simulated data set for dexterous grasping. This captures variations in both observable (e.g. object pose) and unobservable (e.g. surface friction) parameters.

To generate the training set, a simulated depth image of a scene containing a single unfamiliar object is generated. Using either of the generative models GM1 or GM2, grasps are generated and executed in simulation. The success or failure of each simulated grasp is recorded. Producing a good simulation for evaluating grasps is non-trivial. An important problem is that the data set must capture the natural uncertainty in unobservable variables, such as mass and friction. Since many of these parameters are unobservable we are thus creating a data set such that the grasp policy must work across a range of variations. This is thus a form of {\em domain randomisation}. A similar technique has been employed by \cite{mahler2017dex}, but we extend it from a single grasp quality metric to full rigid body simulation.

\subsection{Features and Constraints of the Virtual Environment}
\label{subsection:environment}

\begin{figure}
  \includegraphics[width=\linewidth]{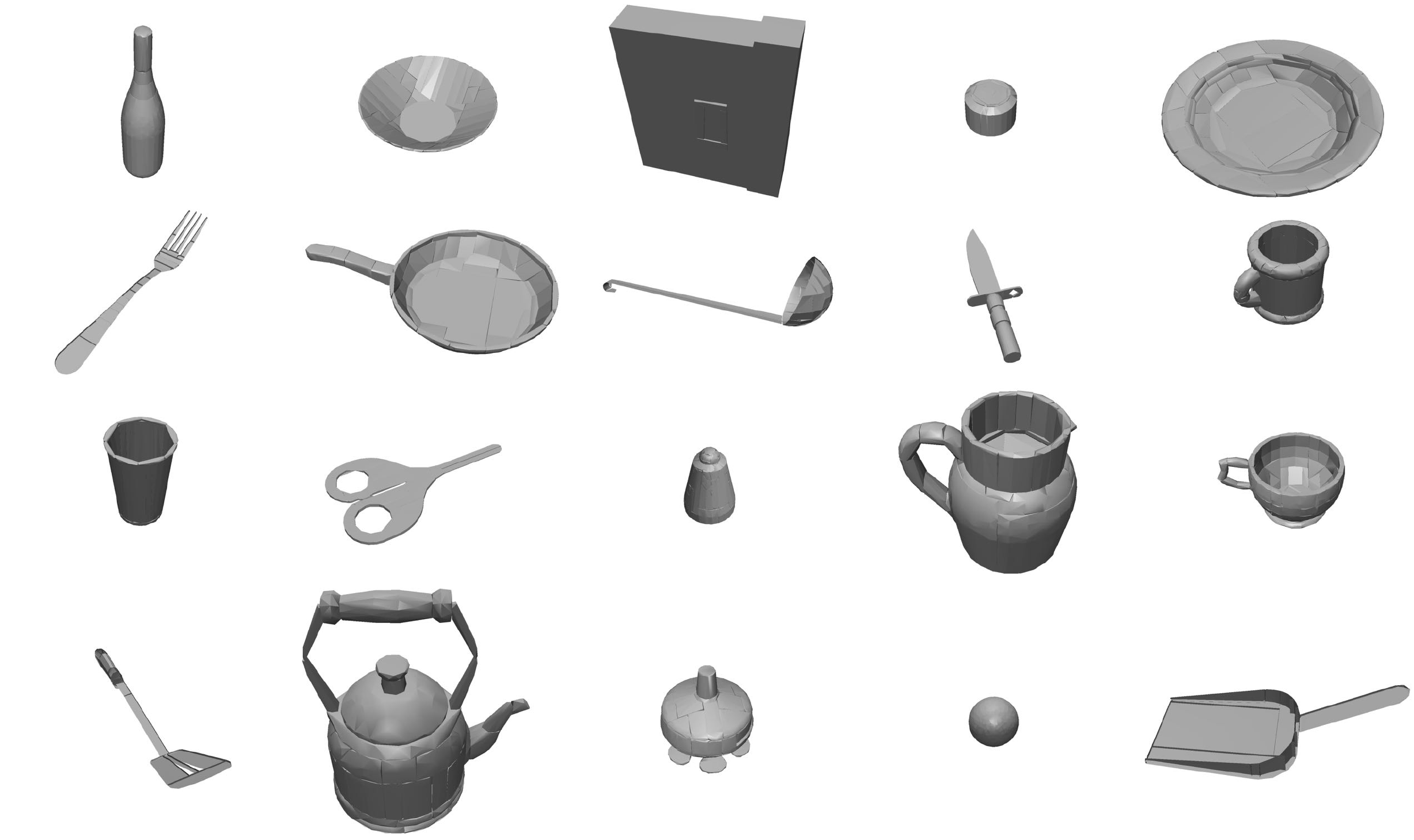}
  \caption{A sample of the 294 objects from all 20 object classes.
  \label{fig:allObjects}}
\end{figure}

The collected 3D model dataset contains 294 objects from 20 classes, namely, bottles, bowls, cans, boxes, cups, mugs, pans, salt and pepper shakers, plates, forks, spoons, spatulas, knives, teapots, teacups, tennis balls, dustpans, scissors, funnels and jugs (Figure \ref{fig:allObjects}). All objects in the dataset can be grasped using the DLR-II hand, although there are limitations on how some object classes can be approached. For example, teapots and jugs are not easy to grasp except by their handles due being larger than the hand's maximum aperture, while small objects such as salt and pepper shakers can be approached in more creative ways. The number of objects in each class varies from 1 (dustpan) to 25 (bottles). Long/thin objects such as kitchen utensils are placed vertically in a short, heavy stand in order to make them graspable without touching the table. This reflects the real-world scenario, as attempting to grasp a spatula lying on a table would be dangerous for the robotic hand. In total, 250 objects from all 20 classes were allocated for training and validation, while the remaining 44 objects from 19 classes belong to the test set.

We employ MuJoCo \cite{MuJoCo} as the rigid-body simulator. Since MuJoCo requires that objects comprise of convex parts, all 294 objects were decomposed into convex parts using V-HACD algorithm \cite{V-HACD}. The number of sub-parts varies from 2 to 120.

During the scene creation, the object is placed on the virtual table at a pseudo-random pose. Most objects are placed in a canonical upright pose, and only randomly rotated around the gravity axis (akin to a turntable). The objects belonging to the mug and cup classes have fully random 3D rotations, as it is possible to grasp them in almost any setting.

To achieve domain randomisation, prior distributions for mass, size and frictional coefficient were estimated from real-world data. The properties of simulated objects are sampled from these priors. For each object its mean size, mass and friction coefficient are matched to a real counterpart. For each trial, the size is randomly scaled by a factor in the range [0.9,1.1], while remaining within the grasp aperture of the hand. Object mass is uniformly sampled from a category specific range, estimated from real objects (Table~\ref{fig:weights}). The friction coefficient of each object is sampled from a range of $[0.5, 1]$ in MuJoCo default units, intended to simulate surfaces from low-friction (metal) to high-friction (rubber). This variation is critical to ensuring that the evaluative model will predict the robustness of a grasp to unobservable variations.
\begin{table}[]
\centering
\caption{Mass ranges for each object class (grams).}
\label{fig:weights}
\resizebox{\linewidth}{!}{\begin{tabular}{|l|l|l|l|l|l|l|}
\hline
Bottle & Bowl     & Box     & Can     & Cup    & Fork    & Pan     \\ \hline
30-70  & 50-400   & 50-500  & 200-400 & 30-330 & 40-80   & 150-450 \\ \hline
Plate  & Scissors & Shaker  & Spatula & Spoon  & Teacup  & Teapot  \\ \hline
40-80  & 50-150   & 100-160 & 40-80   & 40-80  & 150-250 & 500-800 \\ \hline
Jug    & Knife    & Mug     & Funnel  & Ball   & Dustpan &         \\ \hline
80-200 & 50-150   & 250-350 & 40-80   & 50-70  & 100-150 &         \\ \hline
\end{tabular}}
\end{table}
 
For depth image simulation the Carmine 1.09 depth sensor installed on the robot is simulated with a modified version of the Blensor Kinect sensor simulator \cite{KinectSimulator}. For each object, we vary the camera orientation and distance from the object, as well as object mass, friction, scale, location and orientation. We add a small three-dimensional positional noise to each point in the sensor output to simulate calibration errors.

A 3D mesh-model of the DLR-II hand has been used in the simulator. There are no kinematic constraints on how the hand may grasp an object, other than collisions with the table. To ensure realism, we use impedance control for the hand.
\begin{figure}
  \includegraphics[width=\linewidth]{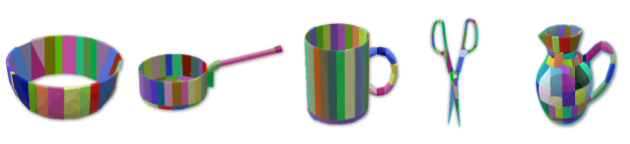}
  \caption{Approximate convex decomposition of some objects in our dataset. Best viewed in colour.}
  \label{fig:objectDecomposition}
\end{figure}

Table \ref{fig:graspperf} shows the success rates of the generated grasps in each class, when attempted with the grasps ranked by the Generative Model (GM1). The sampled grasps perform well on a number of classes including Dustpans, Scissors, Spoons, and Mugs. Some objects can only be grasped in certain ways, i.e. not all 10 training grasps are applicable to all objects.

\begin{table}[]
\centering
\caption{The average and \textbf{top} grasp success rates (\%) of GM1 on simulated data.}
\label{fig:graspperf}
\resizebox{\linewidth}{!}{\begin{tabular}{|l|l|l|l|l|l|l|}
\hline
Bottle & Bowl     & Box     & Can     & Cup    & Fork    & Pan     \\ \hline
35 - \textbf{47} & 26 - \textbf{61}   & 16 - \textbf{30}  & 41 - \textbf{92} & 44 - \textbf{59} & 59 - \textbf{68}   & 37 - \textbf{57} \\ \hline
Plate  & Scissors & Shaker  & Spatula & Spoon  & Teacup  & Teapot  \\ \hline
50 - \textbf{95}  & 62 - \textbf{69}   & 47 - \textbf{53} & 57 - \textbf{65}   & 63 - \textbf{82}  & 48 - \textbf{91} & 26 - \textbf{23} \\ \hline
Jug    & Knife    & Mug     & Funnel  & Ball   & Dustpan &         \\ \hline
24 - \textbf{43} & 58 - \textbf{65}   & 40 - \textbf{80} & 52 - \textbf{65}   & 28 - \textbf{82}  & 60 - \textbf{78} & 45 - \textbf{63} \\ \hline
\end{tabular}}
\end{table}


\subsection{Data Collection Methodology}
\label{subsection:dataCollection}

The data set is divided into units called \textit{scenes}, where each scene comprises a single object placed on a table. This object has a specific set of physical parameters, as described below. Many views and grasps are attempted per scene. Below, we specify the time flow of data collection:

\begin{enumerate}
\item A novel instance of an object from the dataset is generated and placed on a virtual table. Variations are applied to object pose, scale, mass, and friction coefficients.
\item A simulated camera takes a depth image $I_s$ of the scene, converted to a point cloud $P_s$. The viewpoint ${elevation}_s$ of the view point is from 30-57 degrees. The ${azimuth}_s$ is sampled from $[0, 2\pi]$. 
\item All points in the point cloud $P_s$ are shifted by a three-dimensional vector sampled from a Gaussian distribution with parameters $\mu=0$ and $\sigma = 0.004$ (unit: meter).
\item Given $P_s$, the chosen generative model (GM1 or GM2) proposes the candidate grasps. For GM1 and GM2, we choose up to 10 and 50 top grasps per each one of the 10 training grasps, respectively.
\item The grasps are applied to the object in simulation. Before the execution of each grasp, we run a collision check with the virtual table (without the object). The grasps that fail this test are marked as \textit{collided}.
\item 19 further simulated depth images are taken from other viewpoints around the object, as explained in step 2. Images with fewer than 250 depth points are discarded. We then sample with replacement from the remaining images and associate each sampled image and viewpoint with a grasp created in step 3.
\item The grasp outcome, trajectory and depth image are stored for each trial. The grasp parameters are converted to the camera frame for the associated view.
\end{enumerate}


In each scene $S_i$, a number of depth images are taken $\{I_{ik}\}_{k=0}^{20}$, in the manner explained above. The first image $I_{i0}$ is used to generate grasps, as explained in Section \ref{section:generative}. We typically perform 100-500 grasps per scene. Attaching different views to each grasp instead of the seed image $I_{i0}$ ensures there is more variation in terms of viewpoints, resulting in a richer dataset.

Once a grasp is performed in simulation, it is considered a success if an object is lifted one metre above the table, and held there for two seconds. If the object slips from the hand during lifting or holding, the grasp is a failure. 

\begin{figure}[t]
\includegraphics[width=\columnwidth]{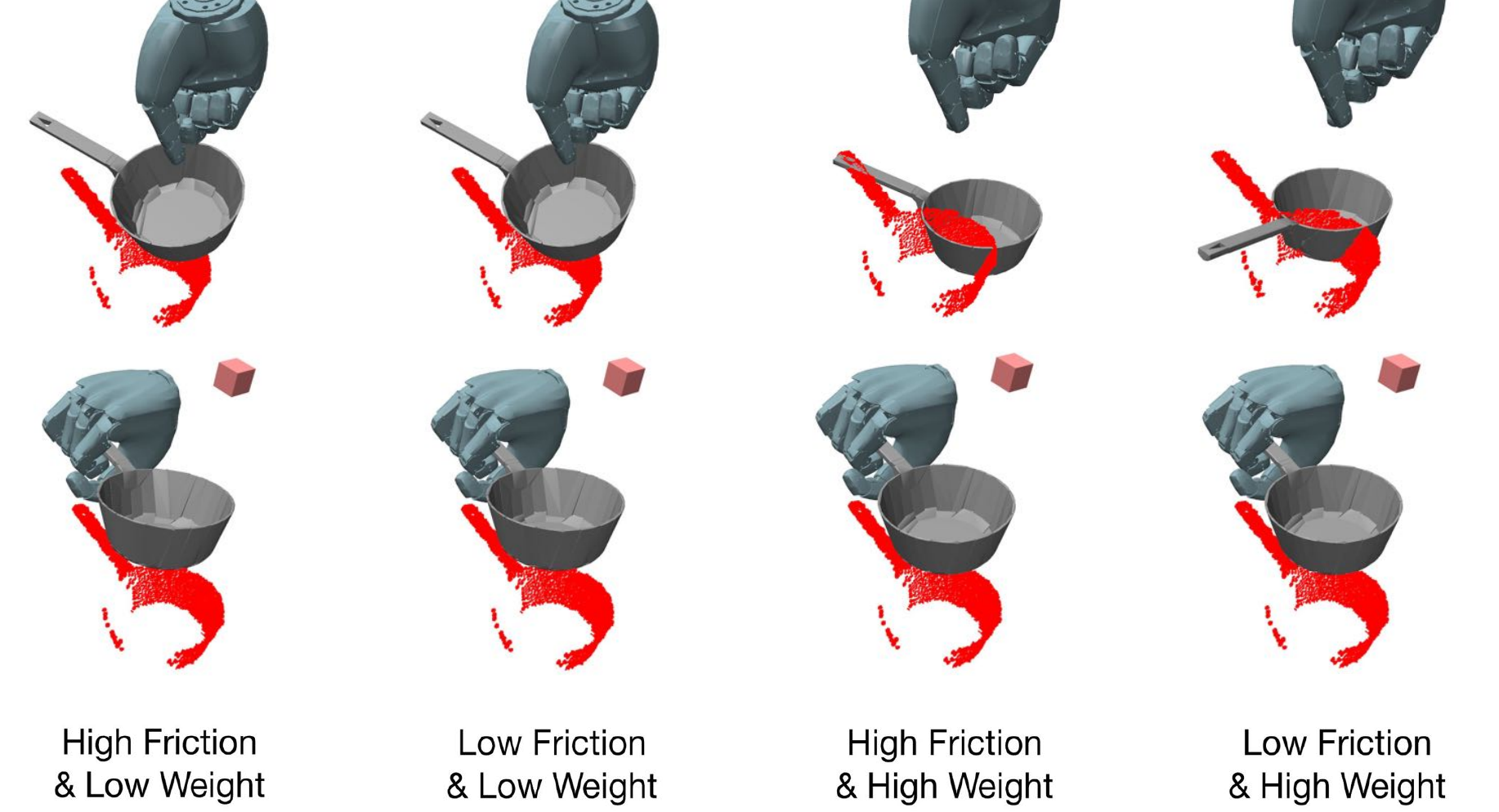}
\caption{Creating a data set for robust evaluation. (Top row) The same pinch grasp, executed on the same object, with varying friction and mass parameters. (Bottom row) A more robust power grasp, executed on the same object, with the same variation in friction and mass. \label{fig:evaluative-training}}
\end{figure}

Using this method, we generated a data set (DS1) of 1.28 million simulated grasps using GM1 as the generative model and a data set of 1.136 million additional grasps (DS2) using GM2 \footnote{Visit \href{https://rusen.github.io/DDG/}{https://rusen.github.io/DDG} to download the data.}. Each grasp in DS1-test and DS2 can be replayed in MuJoCo and the sets are decomposed for train, validation and test purposes. We give the dataset statistics in Table~\ref{tab:data}. The ratio of successful grasps in the dataset is less than 50\% for GM1, and is more than 50\% for GM2. In order to have a balanced training set, DS1 and DS2 only contain scenes that have at least one successful grasp. During training, the datasets were balanced by under-sampling the failure cases in DS1-Tr and over-sampling the failure cases for DS2-Tr. No balancing was performed for the validation and test sets.
\begin{table*}[t]
\centering
\caption{Statistics of the simulated data sets.}
\label{tab:data}
\begin{tabular}{|l|l|l|l|l|l|l|l|l|l|l|} \hline
Data set & Generative &  Subset & \# Scenes & Top-grasp & Top-grasp & Top grasp & Total & Total  & Total  & Total \\ 
              & Model         &              &                   &  \# succs  & \# fails       & \% succs  & grasps   & \# succs      & \# fails  & \% succs  \\ \hline
 DS1-Tr & GM1 & Train & 17714 & 10100 & 7614 & 57.0\% & 1,058,430 & 479,941 & 578,489 & 45.3\% \\ \hline
 DS1-V  & GM1 & Validate & 2309 & 1290 & 1019 & 55.9\% & 122,944 & 61,256 & 61,688 & 49,8\% \\ \hline
 DS1-Te & GM1& Test & 1539 & 1070 & 469 & 69.5\% & 99,521 & 48,084 & 51,437 & 48.3\% \\ \hline
 DS2-Tr  & GM2 & Train & 5377 & 3771 & 1606 & 70.1\% & 943,481 & 533,282 & 410,199 & 56.5\% \\ \hline
 DS2-V   & GM2 & Validate & 544 & 378 & 166 & 69.4\% & 68,586 & 39,559 & 29,027 & 57.7\% \\ \hline
 DS2-Te  & GM2 & Test & 988 & 781 & 207 & 79.0\% & 124,137 & 73,836 & 50,301 & 59.5\% \\ \hline
\end{tabular}
\end{table*}

\section{The Generative Evaluative Architecture} \label{section:evaluative}
The grasping system proposed, shown in Figure \ref{fig:systemArchitecture}, consists of a learned generative model and an evaluative model. The generative model is a method that generates a number of candidate grasps given a point cloud, as explained in the previous section. An evaluative model is paired with a generative model in order to estimate a probability of success for each candidate grasp. All evaluative models process the visual data and hand trajectory parameters in separate pathways, and combine them to feed into a third processing block to produce the final success probability. In addition, we present techniques for grasp optimisation using the EM as the objective function, using both Gradient Ascent (GA) and Simulated Annealing (SA). Finally, we may train each model with either the data set of simulated grasps generated by GM1, by GM2, or both. Table \ref{table:GEBreakdown} shows a the full list of 17 variants we test.

\begin{table}[]
\centering
\begin{tabular}{|l|l|l|l|l|l|l|}
\hline
Variant & GM/  & EM & Opt'  & Training Set \\ 
 & Testset & & Meth' & \\ \hline
V1 & GM1    & - & - & 10 grasps  \\ \hline
V2 & GM2    & - & - & 10 grasps  \\ \hline
V3 & GM1/DS1-Te & EM1 & - & DS1-Tr \\ \hline
V4 & GM1/DS1-Te & EM2 & - & DS1-Tr \\ \hline
V5 & GM1/DS1-Te & EM3 & - & DS1-Tr  \\ \hline
V6 & GM1/DS1-Te & EM1 & - & DS1-Tr + DS2-Tr \\ \hline
V7 & GM1/DS1-Te & EM2 & - & DS1-Tr + DS2-Tr \\ \hline
V8 & GM1/DS1-Te & EM3 & - & DS1-Tr + DS2-Tr \\ \hline
V9 & GM2/DS2-Te & EM1 & - & DS1-Tr + DS2-Tr \\ \hline
V10 & GM2/DS2-Te & EM2 & - & DS1-Tr + DS2-Tr \\ \hline
V11 & GM2/DS2-Te & EM3 & - & DS1-Tr + DS2-Tr \\ \hline
V12 & GM1/DS1-Te & EM3 & GA1 & DS1-Tr + DS2-Tr \\ \hline
V13 & GM1/DS1-Te & EM3 & GA2 & DS1-Tr + DS2-Tr \\ \hline
V14 & GM1/DS1-Te & EM3 & GA3 & DS1-Tr + DS2-Tr \\ \hline
V15 & GM1/DS1-Te & EM3 & SA1 & DS1-Tr + DS2-Tr \\ \hline
V16 & GM1/DS1-Te & EM3 & SA2 & DS1-Tr + DS2-Tr \\ \hline
V17 & GM1/DS1-Te & EM3 & SA3 & DS1-Tr + DS2-Tr \\ \hline
\end{tabular}
\caption{The evaluated combinations of architecture, generative model/test set, training set, and optimisation method (Gradient Ascent (GA) or Stochastic Simulated Annealing (SA).}
\label{table:GEBreakdown}
\end{table}

In this section, the three proposed evaluative model (EM) architectures are explained. The grasp generator models, GM1 and GM2, given in the previous section, require very little training data to train, here being trained from 10 example grasps. 
These generative models do not, however, estimate a probability of success for the generated grasps. An evaluative model, which is a Deep Neural Network (DNN), is used specifically for this purpose. DNNs have shown good performance in learning to evaluate grasps using grippers \cite{levine16,lenz2015deep}. They have also been applied to generating pre-grasps, so as to perform power grasps with dexterous hands \cite{varley2015generating,lu2017planning}.

\begin{figure*}[t]
\centering
\subfloat[Evaluative Model 1]{%
  \includegraphics[width=0.75\textwidth]{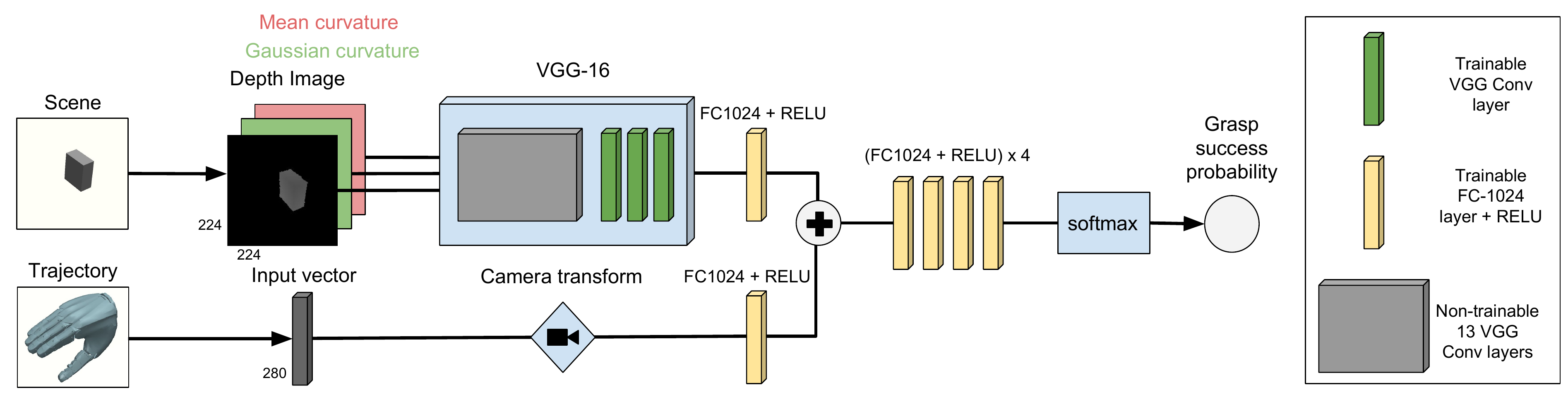}
}

\subfloat[Evaluative Model 2]{%
    \includegraphics[width=0.66\textwidth]{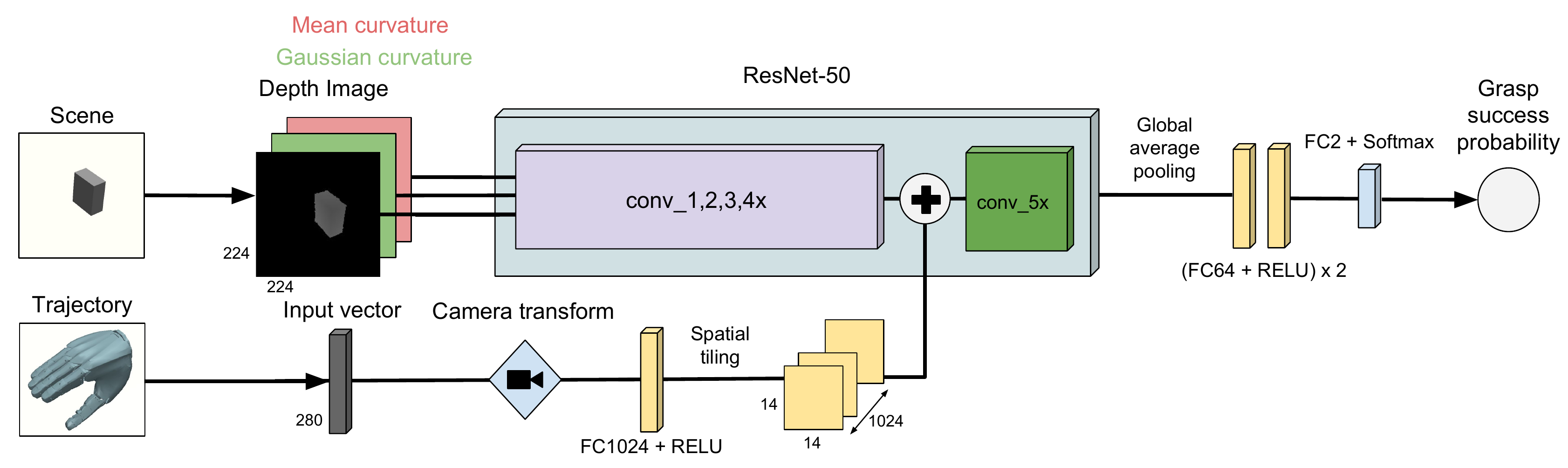}
}

\subfloat[Evaluative Model 3]{%
    \includegraphics[width=0.66\textwidth]{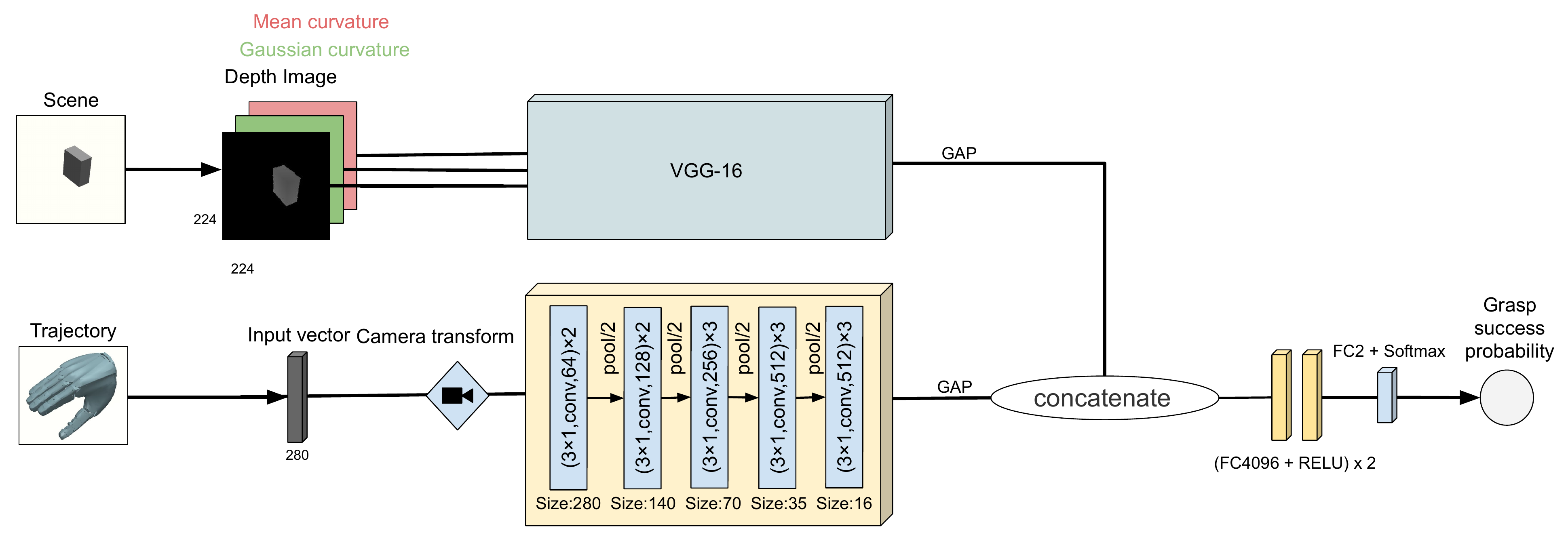}
}

\caption{The three proposed evaluative network architectures. Similar to \cite{Levine1}, the two channels of information (visual data and grasp parameters) are processed in parallel and combined to reach the final decision. RELU activations are used throughout the models, except for the final softmax layers. A final softmax layer has grasp success and and failure nodes, and learns to predict the success probability of a grasp. (a) EM1, a VGG-16 based model, where the first 13 layers of VGG-16 are frozen. (b) EM2, a ResNet-50-based \cite{HeZRS15} network. First four blocks are used for feature extraction, and the rest of the network is used to learn joint features. (c) Second model based on VGG-16. In EM3, the channels are joined via concatenation, not addition.
\label{fig:networkArchitecture2}}
\end{figure*}

We tested three evaluative models. The first is based on the VGG-16 network \cite{Simonyan14c}, named Evaluative Model 1 (EM1), and shown in Figure \ref{fig:networkArchitecture2} (a). A version based on the ResNet-50 network, termed EM2, is shown in Figure \ref{fig:networkArchitecture2} (b). Finally, EM3 (Figure \ref{fig:networkArchitecture2} (c)) is also based on VGG-16. All EMs are initialised with ImageNet weights. Regardless of the type, an EM has the functional form $f(I_t, h_t)$, where $I_t$ is a colourised depth image of the object, and $h_t$ contains a series of wrist poses and joint configurations for the hand, converted to the camera's frame of reference. The network's output layer calculates a probability of success for the image-grasp pair $I_t$, $h_t$. The model processes the grasp parameters and visual information in separate channels, and combines them to feed into a feedforward pipeline that produces the output.


The depth image is colourised before it is passed as input to the evaluative network. This converts the 1-channel depth data to a 3-channel RGB image. We first crop the middle $460 \times 460$ section of the $640 \times 480$ depth image, and down-sample it to $224 \times 224$. Two more channels of the same dimension are added corresponding to the mean and Gaussian curvatures. 
This procedure both provides meaningful depth features to the network, and makes the input compatible with VGG-16 and ResNet, which require images of size $224 \times 224 \times 3$.

The grasp parameter data $h_t$ consists of 10 trajectory waypoints represented by $27 \times 10 = 270$ floating point numbers, and 10 extra numbers reserved for the grasp type. Each of the 10 training grasps is treated as a different class, and $h_t$ uses the 1-of-N encoding system. Based on the grasp type ([1-10]), the corresponding entry is set to 1, while the rest remain 0. The grasp parameters are converted to the coordinate system of the camera which was used to obtain the corresponding depth image. In EM1 and EM2, the parameters are processed with a fully-connected (FC-1024) layer, and the output is \textit{element-wise added} to the visual features, while EM3 uses a convolutional approach. In all networks, the joint visual features and grasp parameter data are joined in higher layers.

All FC layers have RELU activation functions, except for the output layer, which uses 2-way softmax in all EM variants. The output layer has two nodes, corresponding to the success and failure probabilities of the grasp. A cross-entropy loss is used to train the neural network, as given in \eq\ref{equation:crossentropy}.

\begin{equation}
H_{y'}(y) := - \sum_{i} ({y_i' \log(y_i) + (1-y_i') \log (1-y_i)})
\label{equation:crossentropy}
\end{equation}
where $y_i'$ is the class label of the grasp, which is either 1 (success) or 0 (failure), and $y_i = f(I_i, h_i)$ is is the predicted label of the grasp pair ($I_i$, $h_i)$.

The individual models are now introduced below. Only their unique properties are highlighted.

\subsection{Evaluative Model 1 (EM1)}

Figure~\ref{fig:networkArchitecture2} (a) shows the architecture of the first proposed evaluative network. The colourised depth image is processed with the VGG-16 network \cite{Simonyan14c} to obtain the image features. We froze the first 13 layers in order to reduce overfitting.

The grasp parameters and image features pass through two FC-1024 layers in order to obtain two feature vectors of length 1024. The features are combined using the element-wise addition operation, and fed into 4 FC-1024 layers. Similarly with \cite{Levine1}, we use addition, not concatenation. This follows the observation that addition yielded a marginally better performance in the experiments. Furthermore, concatenation and addition can be considered as interchangeable operations in this context \cite{dumoulin2018feature-wise}. The final FC-1024 layers form the associations between the visual features and hand parameters, and contain most of the trainable parameters in the network.

\subsection{Evaluative Model 2 (EM2)}


EM2 (Figure~\ref{fig:networkArchitecture2} (b)) uses the ResNet-50 architecture in order to obtain the image features. In the EM2 architecture, ResNet-50 network is broken down into two parts: the first 4 convolutional blocks are used to extract the visual features. The final block, which has 9 randomly-initialised convolutional layers, combines the image features and grasp parameters. Similarly with EM1, element-wise addition joins the two channels of information. Spatial tiling is used to convert the processed grasp parameters, a vector of size $1024$, to a matrix of size $14 \times 14 \times 1024$. Because the last block processes combined information, EM2 is designed with only 2 FC-64 layers.

\subsection{Evaluative Model 3 (EM3)}

This model, as for EM1 (Figure~\ref{fig:networkArchitecture2} (c)), uses VGG-16 as the visual backbone. All 16 layers of VGG-16 are trained. The hand trajectory parameters pass through a feature extraction network before being concatenated with the visual features. The combined part of the network contains two high-capacity FC-4096 layers, followed by a FC2+softmax layer.

EM3, in contrast to EM1 and EM2, uses convolutional layers for processing input grasp trajectories. The trajectory sub-network is similar to VGG-16 in that it contains 5 blocks, comprising 13 convolutional layers. The convolutional filters have a width of 3. The sizes under the blocks are input dimensions. Global Average Pooling (GAP) is performed to obtain 512 features coming from both sides, which are concatenated and run through two FC-4096 layers.

All models were trained and tested on simulated data. EM2 and EM3 were tested on the real robot setup. 

\begin{table*}[t]
\centering
\begin{tabular}{|l|l|l|l|l|l|l|l|l|l|l|}
\hline
Variant \# & \multicolumn{2}{|c|}{Selected grasp} &  Succ \% & Fails as \% & Test set & \multicolumn{5}{|c|}{Prediction Performance} \\ \cline{2-3} \cline{7-11}
 & Succs & Fails &  & of V1 fails & / GM & TP & FP & TN & FN & Accuracy \\ \hline
V1 & 1070   & 469 & 69.53\% & 100\% & GM1 & - & - & - & - & - \\ \hline
V2 &  781   & 207 & 79.05\% & 68.7\% & GM2 & - & - & - & - & - \\ \hline
V3 & 1352 & 187 & 87.85\% & 39.9\% & GM1 & 37840 &	12226 & 39211 & 10244 & 77.42\% \\ \hline
V4 &  1361 & 178 & 88.43\% & 38.0\% & GM1 & 40234 &	14475 & 36962 & 7850 & 77.57\% \\ \hline

V5 & 1361 & 178 & 88.43\% & 38.0\%& GM1 & 39603 & 14122 &37315 & 8481	& 77.29\% \\ \hline

V6 & 1375 & 164 & 89.34\% & 35.0\% & GM1 & 37584 &	11514 & 39923 &10500	& 77.88\% \\ \hline
V7 & 1363 & 176 & 88.56\% & 37.5\% & GM1 & 39332 &	12020 & 39417 & 8752 & 79.13\% \\ \hline
V8 & 1378 & 161 & 89.54\% & 34.3\% & GM1 & 37832 &	11361 & 40076	& 10252 & 78.28\% \\ \hline
V9 & 887 & 101 & 89.78\% & 33.5\% & GM2 & 61866 &	11454& 38847& 11970 & 81.13\% \\ \hline
V10 & 893 & 95 & 90.38\% & 31.6\% & GM2 & 64309 &	12517 & 37784 & 9527 & 82.24\% \\ \hline
V11 & 894 & 94 & 90.49\% & 31.2\% & GM2 & 61611 & 9792 & 40509 & 12225 & 82.26\% \\ \hline
V12 & 1319 & 220 & 85.71\% & 47.0\% & GM1 & - & - & - & - & - \\ \hline
V13 & 1375 & 164 & 89.34\% & 35.0\% & GM1 & - & - & - & - & - \\ \hline
V14 & 1366 & 173 & 88.76\% & 37.0\% & GM1 & - & - & - & - & - \\ \hline
V15 & 1153 & 386 & 74.92\% & 82.0\% & GM1 & - & - & - & - & - \\ \hline
V16 & 1377 & 162 & 89.47\% & 35.0\% & GM1 & - & - & - & - & - \\ \hline
V17 & 1163 & 376 & 75.57\% & 80.0\% & GM1 & - & - & - & - & - \\ \hline
\end{tabular}
\caption{Simulation results for all variants tested.}
\label{table:Results-sim}
\end{table*}

\subsection{EM training methodology}
Variants V3-V5 were trained using DS1-Tr \footnote{10\% of DS1-Tr failure cases are sampled from the grasps that collide with the table, and we preserved the colliding grasps in DS1-V. This was done to ensure EMs do not propose such grasps in real robot experiments.}. Variants V6-V17 were trained using the combined data set from DS1-Tr and DS2-Tr \footnote{The grasps that collide with the table were removed from DS2. Filtering became unnecessary since the overall quality of grasps by GM2 is better.}. The Gradient Descent(GD) optimiser was employed with starting learning rate of 0.01, a dropout rate of 0.5, and early stopping. We halve the learning rate every 5 epochs during training.

\subsection{Grasp optimisation using the EM}

So far we have considered only Generative-Evaluative architectures where the Evaluative Model merely ranks the grasp proposals. As proposed by Lu et al. \cite{lu2017planning} we may also use the EM to improve grasp proposals. This boils down to searching the grasp space driven by the EM as the objective function. This may be by gradient ascent or simulated annealing. The methods V12-17 use V8 as the objective function, hence V8 should be treated as the baseline. We employed both gradient based optimisation and simulated annealing.

\subsubsection{Gradient based optimisation}
Lu et al. \cite{lu2017planning} proposed gradient ascent (GA), modifying the grasp parameters input to the EM with respect to the output predicted success probability. They initialised with a heuristically selected pre-grasp. We initialise with the highest ranked grasp according to the EM. We investigated three variants:
\begin{itemize}
\item GA1: Shifts the position of the all waypoints in the grasp trajectory equally. The gradient is the average position gradient across all 10 waypoints.
\item GA2: Tunes the hand configuration by tuning the angle of each finger joint. Every finger joint at each waypoint is treated independently.
\item GA3: Performs GA1 and GA2 simultaneously.
\end{itemize}

\subsubsection{Simulated annealing based optimisation}
Gradient based optimisation is sensitive to the quality of gradient estimates derived from the model. Simulated annealing (SA) based optimisation is more robust to such noise. Therefore, three optimisation routines were implemented using SA:
\begin{itemize}
\item SA1: Shifts the positions of the all waypoints in the grasp trajectory equally. Moves are drawn from a three-dimensional Gaussian with $\mu=0$ and $\sigma=0.001$. 
\item SA2: Scales the angles of the finger joints in the final grasp pose with a single scaling parameter drawn from a Gaussian with $\mu=1$ and $\sigma=0.001$. The initial finger joint angles remain fixed and joint angles of the intermediate waypoints are linearly interpolated. 
\item SA3: Performs SA1 and SA2 simultaneously.
\end{itemize}

\section{Simulation Analysis}
\label{section:simulationAnalysis}
This section presents a simulation analysis of the various architectures based on the two data sets. 
We assess each variant in two different ways. First, for any method with an evaluative model we  measure the prediction accuracy of the EM. We compare the actual outcomes in a test set with the EM's prediction as to whether it is more likely to succeed or fail (output set at a threshold of 0.5). This gives us a confusion matrix from which we can calculate sensitivity, specificity and F1 score. Second, since a robot can only execute one grasp, we can measure the proportion of successful top-ranked grasps for any method. In each analysis the test set effectively replaces the GM as it contains, for any scene, a complete list of grasps. Thus TS1 contains grasps proposed by GM1 and TS2 contains grasps proposed by GM2. This allows us to simulate the effect of different generative models on performance.

\begin{figure}
\centering
\includegraphics[width=0.8\columnwidth]{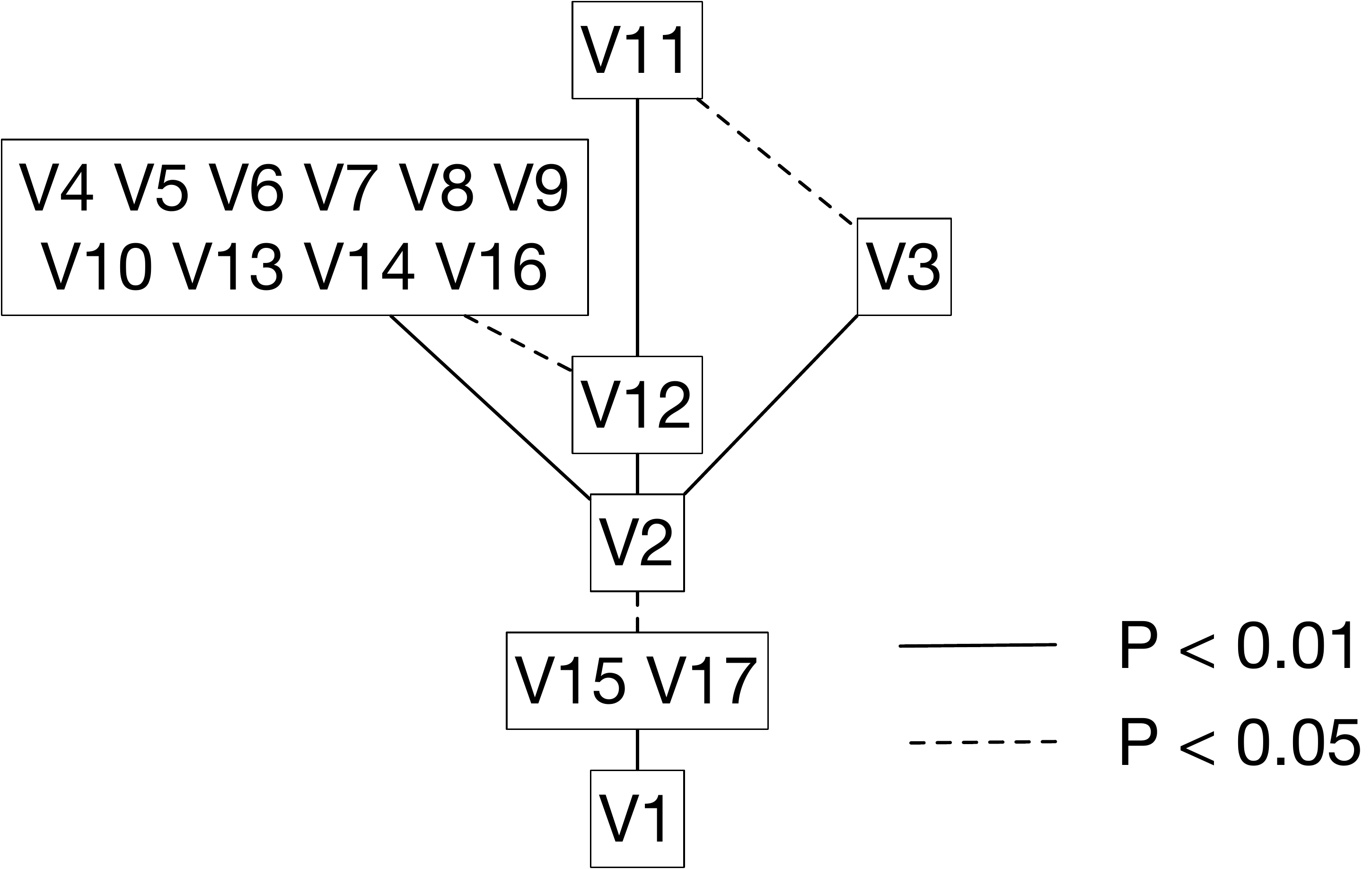}
\caption{Partial order dominance diagram for simulation experiments. We used Fisher's exact test.}
\label{fig:dominance}
\end{figure}

We performed both analyses and the results are given in Table~\ref{table:Results-sim}. A partial order dominance diagram, showing which differences in grasp success rate on the test set are statistically significant using Fisher's exact test, is given in Figure~\ref{fig:dominance}. When assessing pure GM architectures, we can only measure the top ranked grasp success, since the GMs give a grasp likelihood according to the generative model, not a probability of success.

For variants V12-V14, the gradient based optimisation ran for 50 iterations, using a learning rate of 0.001 for position inputs and 0.01 for finger joints.\footnote{We used different learning rates since the parameters are in different units: position in meters and finger joint angles in radians.} For variants V15-V17 the simulated annealing procedure ran for 5 iterations, with 20 random perturbations in each step. We start with a temperature of 0.2 and halve it after every iteration. If the solution does not improve after three steps, optimisation stops. Perturbations that will result in a collision with the table are rejected.

The main findings are as follows. First, of the pure generative models GM2 outperforms GM1, with top ranked grasp successes of 79.05\% and 69.53\% respectively. Second, the joint architectures all outperform both pure GM architectures, starting at 87.85\% of grasps succeeding (V3 based on proposals from GM1 and evaluation by EM1 trained on TS1). Third, the increase in training set size (adding GM2 to GM1) yields a further improvement. We can best measure this by considering the residual number of top grasps that fail as a percentage of the baseline (GM1). On this measure adding the additional data (variants V6-V9) improves performance (over variants V3-V5) by an average of 3\%. 

The results above use GM1 as the generative model. We can measure the benefit of substituting this by GM2. This yields a further reduction in residual failures over GM1 under the same conditions (training with DS1-Tr and DS2-Tr) of 3.5\%.

For both the gradient and simulated-annealing based optimisations, while the predicted probability of success according to the EM rises, the actual success rate in simulation declines for all variants V12-V17. We observed that wrist position changes have a greater negative impact than finger joint. The results suggest that optimising dexterous grasps by the EM is non-trivial. It should be noted that the performance of the gradient ascent was much better than the simulated annealing.

\begin{figure}[t]
\centering
\subfloat[GM1 Ranking Comparison]{%
  \includegraphics[width=0.9\columnwidth]{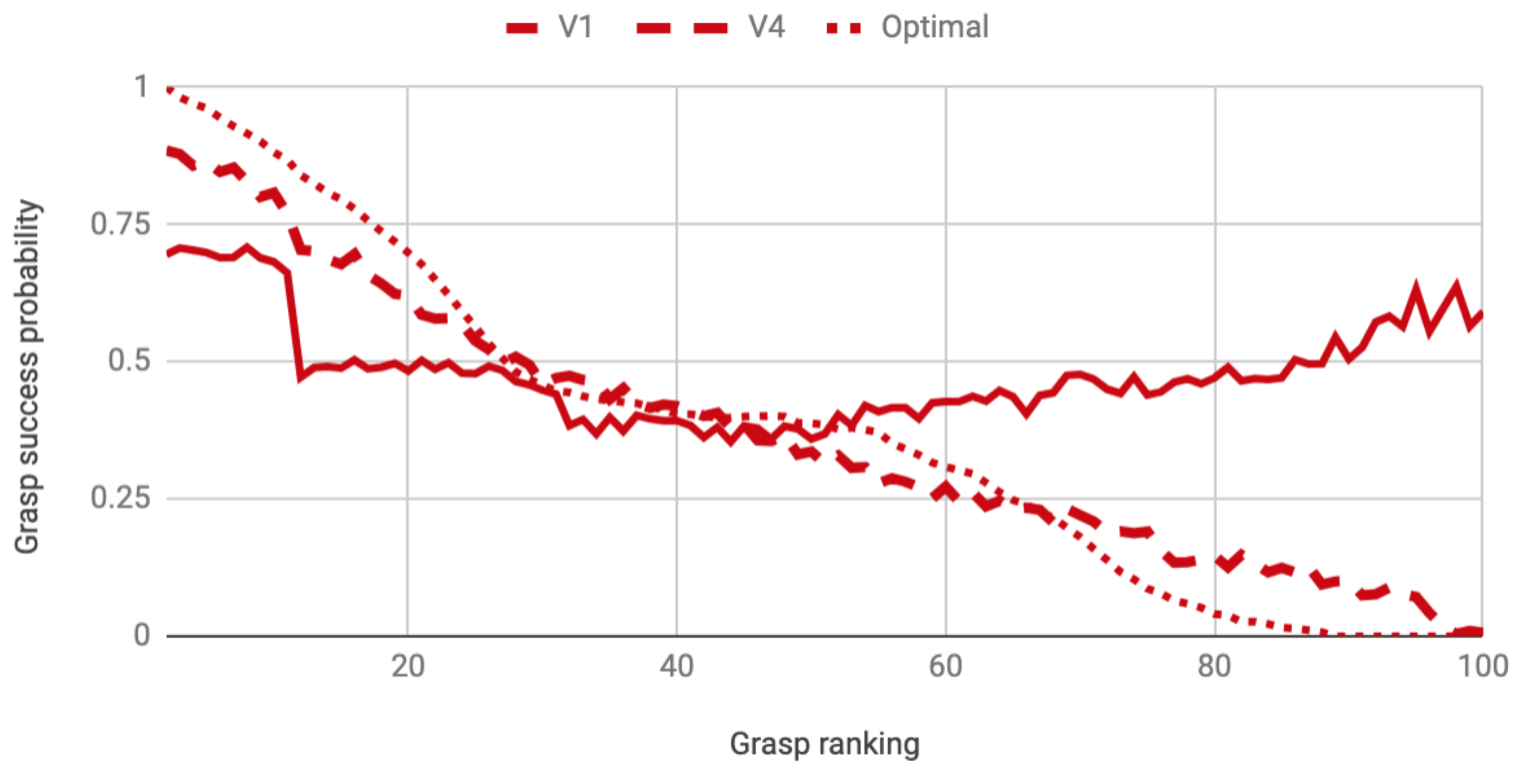}
}

\subfloat[GM2 Ranking Comparison]{%
  \includegraphics[width=0.9\columnwidth]{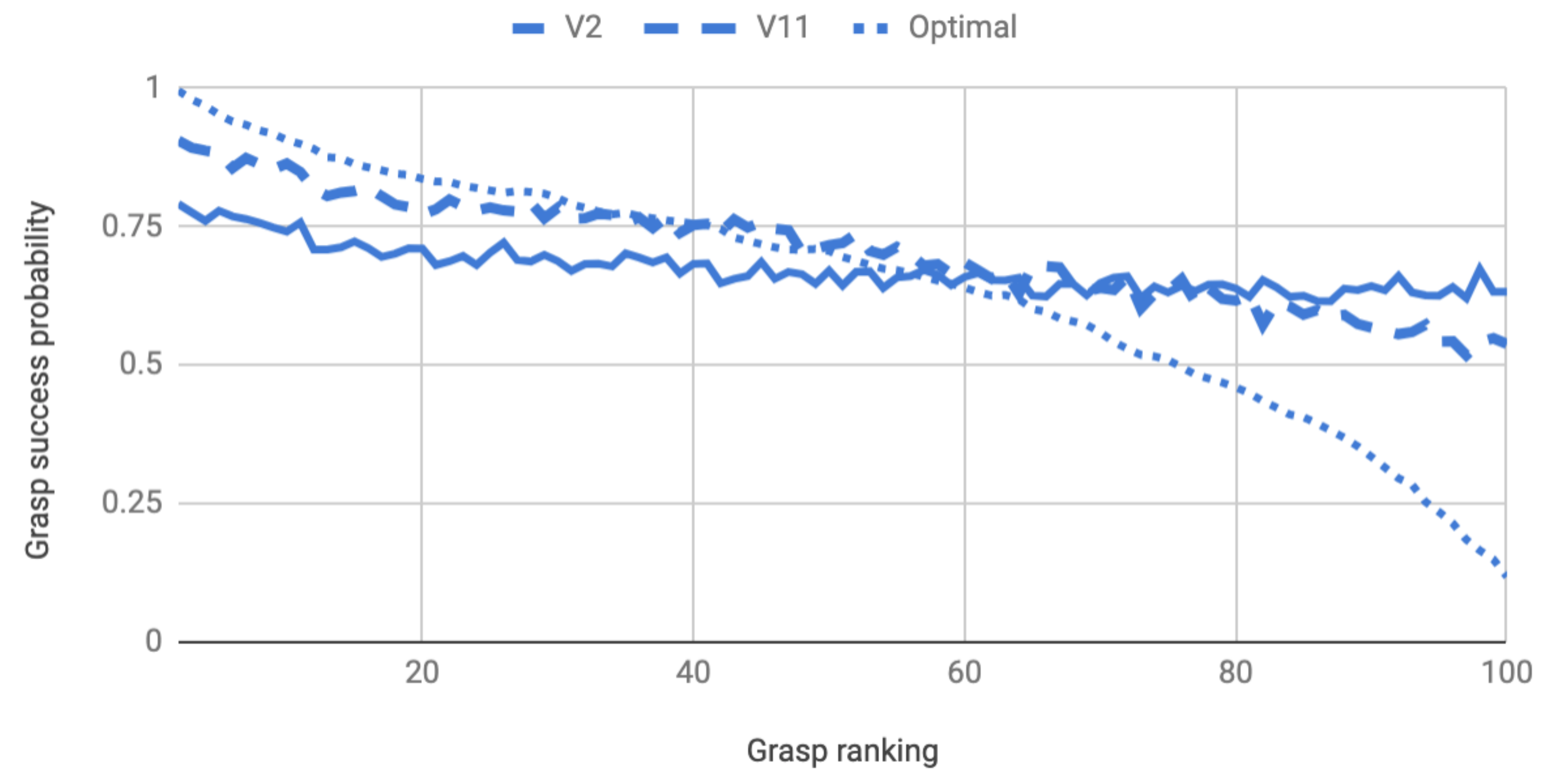}
}
  \caption{Grasp success probability (in simulation) vs. grasp ranking. (a) GM1 vs. V4 vs. optimal ranking (empirical limit). (b) GM2 vs. V11 vs. optimal.
  \label{fig:successvsranking}}
\end{figure}

It is instructive to understand the effect of re-ranking with the EM by referring to Figure~\ref{fig:successvsranking}. This shows the average grasp success probability (across the test set) in simulation against the grasp rank. We observed that the evaluative models are much more effective than the generative models at correctly ranking the grasps. The optimal ranking is also shown. It can be seen that the GEA architectures remove more than half the residual grasp failures by re-ranking so that a good grasp is the first ranked grasp.

In summary, simulation results provide evidence that: (i) pure GM2 outperforms GM1; (ii) adding training data (DS2-Tr to DS1-Tr) improves results; (iii) using GM2 as the generative model in the generative-evaluative architecture improves results; and (iv) that post-rank tuning of the grasp using the EM output as the objective function doesn't improve results.

\section{Real robot experiment}
\label{section:experiments}

\begin{figure}[t]
\begin{center}
  \includegraphics[width=0.9\linewidth]{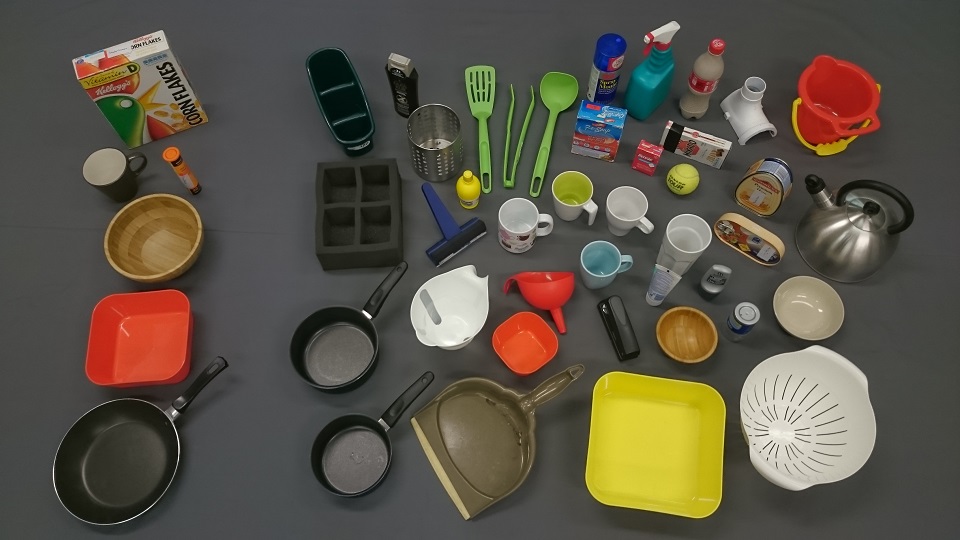}
  \end{center}
  \caption{The real objects. The training objects are on the left, testing objects are on the right.}
  \label{fig:real-objects}
\end{figure}

\begin{table}[b]
\begin{center}
\caption{Performance on the real robot. \label{tab:robot-results}}
\begin{tabular}{|c|c|c|c|c|c|} \hline
Alg & \# succ & \% succ & Alg & \# succ & \% succ \\ \hline
V1  &  28 & 57.1\% & V4   & 37  & 75.5\% \\
V2  & 40 & 81.6\% & V11 & 43  & 87.8\% \\
\hline
\end{tabular}
\end{center}
\end{table}

\begin{figure}[t]
\begin{center}
\includegraphics[width=0.9\columnwidth]{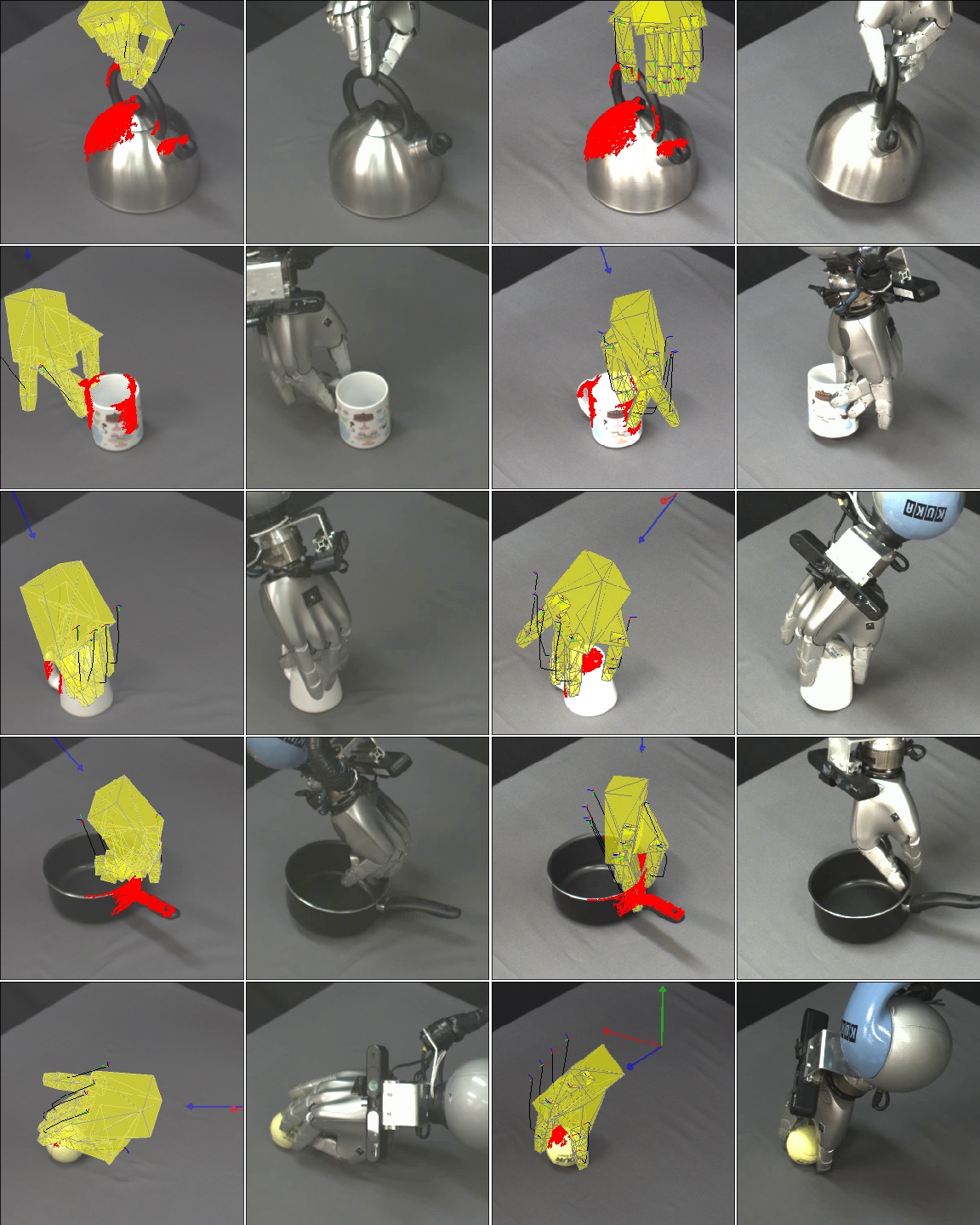}
\caption{V2 vs V11. This shows grasps from methods based on generative model GM2. The V2 grasps are shown in columns 1-2. The corresponding V11 grasps are shown in columns 3-4. These are the cases where V2 failed and V11 succeeded. \label{fig:v2fv11s}}
\end{center}
\end{figure}

\begin{figure}[t]
\begin{center}
\includegraphics[width=0.9\columnwidth]{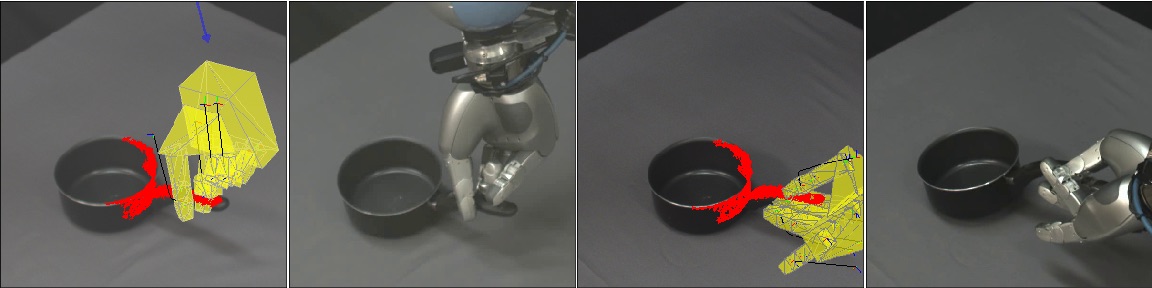}\\
\includegraphics[width=0.9\columnwidth]{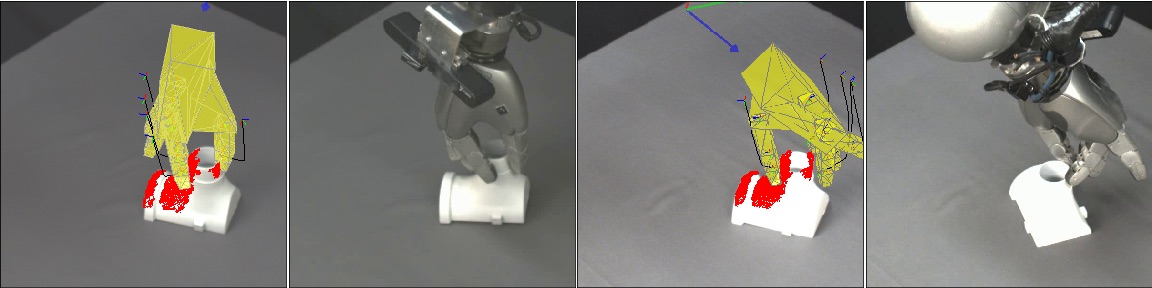}
\caption{V2 vs V11. This shows grasps from methods based on generative model GM2. The V2 grasps are shown in columns 1-2. The corresponding V11 grasps are shown in columns 3-4. The top row shows the case where both failed. The bottom row shows the case where V2 succeeded and V11 failed. \label{fig:v2fsv11f}}
\end{center}
\end{figure}

We compared four variants on the real robot: V1, V2, V4 and V11. V1 and V2 are the pure generative models. V4 is, in simulation, the equal best generative-evaluative method using GM1 as the generative model. It uses EM2 as the evaluative model. V11 is, in simulation, the best performing generative-evaluative method using GM2 as the generative model. This selection allows us to compare the best generative-evaluative methods with their counterpart pure generative models. 

We employed the same real objects as described in \cite{kopicki2019ijrr}. This used 40 novel test objects (Figure~\ref{fig:real-objects}). Object-pose combinations were chosen to reduce the typical surface recovery. Some objects were employed in several poses, yielding 49 object-pose pairs. From the 40 objects, 35 belonged to object classes in the simulation dataset, while the remaining five did not. 

Using this data-set, all algorithms were evaluated on the real-robot using a paired trials methodology. Each was presented with the same object-pose combinations. Each variant generated a ranked list of grasps, and the highest ranked grasp was executed. The highest-ranked grasp based on the predicted success probability of an evaluative network is performed on each scene. A grasp was deemed successful if, when lifted for five seconds, the object then remained stable in the hand for a further five seconds. 

The results are shown in Table~\ref{tab:robot-results}. In each case, the generative-evaluative variant outperforms the equivalent pure GM variant. So that V4 outperforms V1 by 75.5\% grasp success rate to 57.1\% and V11 outperforms V2 87.8\% to 81.6\%. The differences between V11:V1 and V2:V1 are highly statistically significant ($p<0.01$) using McNemar's test. Thus, we have strong support for our main hypothesis, which is that a Generative-Evaluative architecture outperforms a pure generative model. Six of the available grasp types were deployed (pinch support, pinch, pinchbottom, rimside, rim and power edge), showing that a variety of grasps is utilised.



\begin{figure*}
\begin{center}
\includegraphics[width=0.45\textwidth]{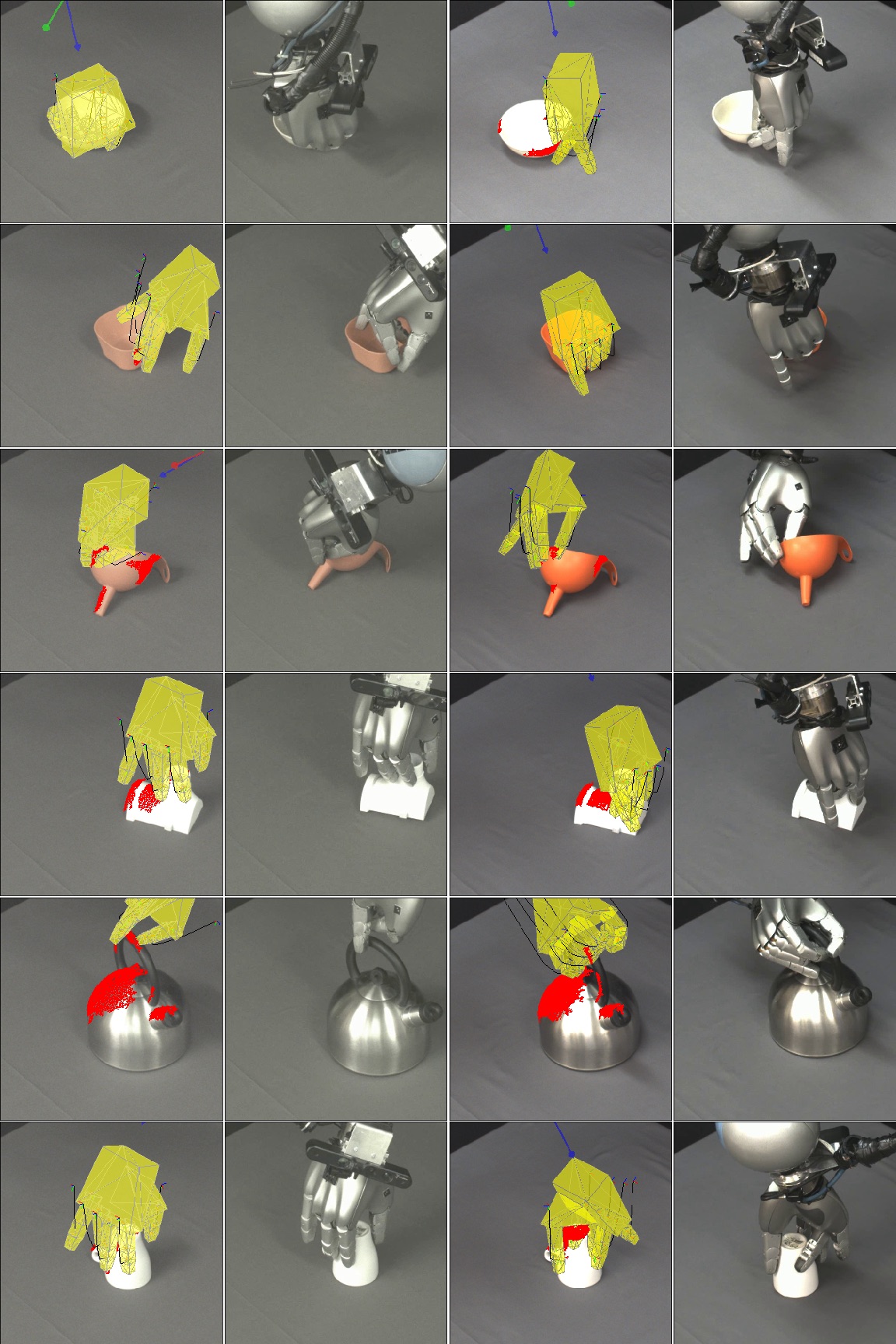}~
\includegraphics[width=0.45\textwidth]{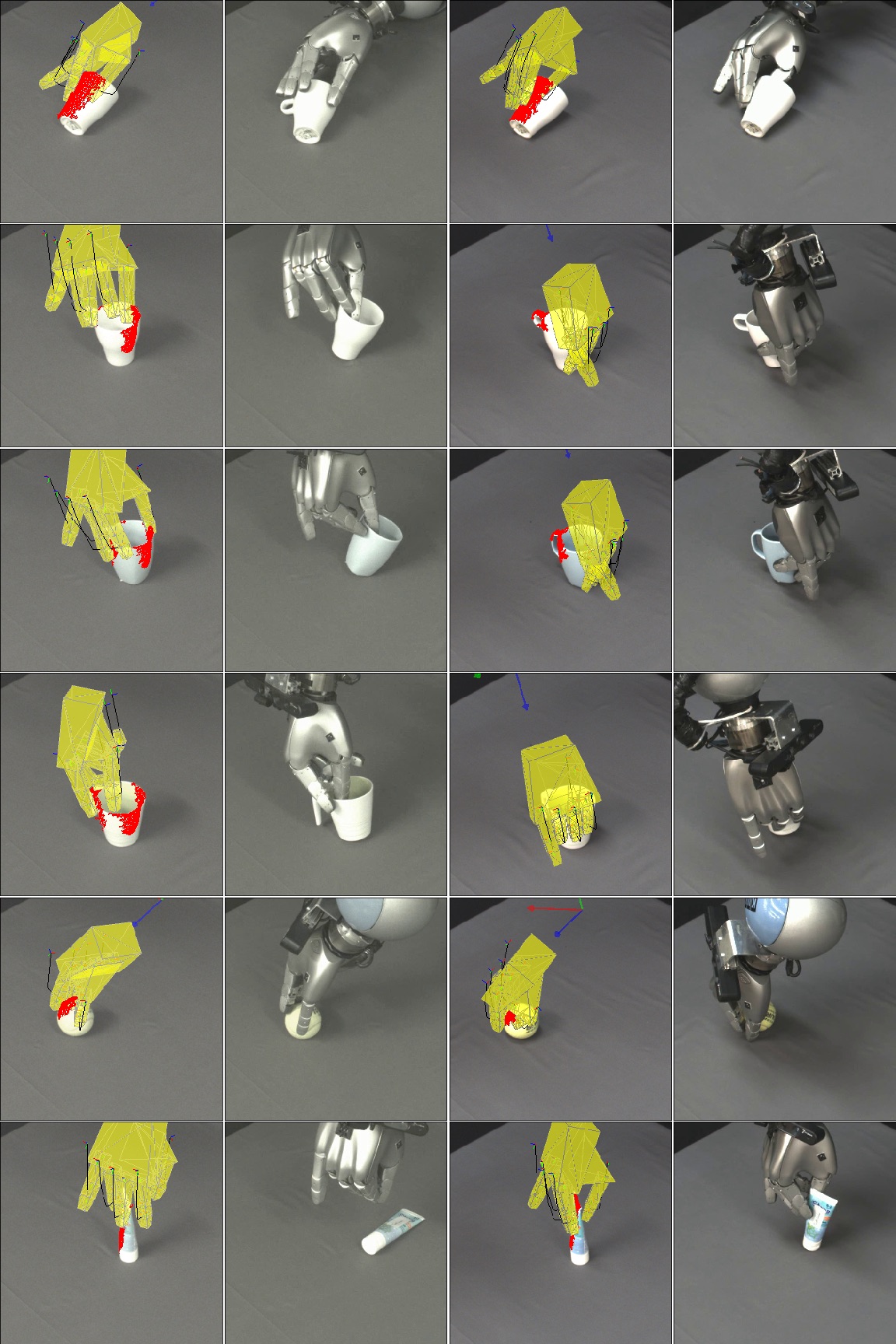}
\caption{V1 vs V4. This shows grasps from methods based on generative model GM1. The V1 grasps are shown in columns 1-2 and 5-6. The corresponding V4 grasps are in columns 3-4 and 7-8. These are the cases where V1 failed and V4 succeeded.\label{fig:v1fv4s}}
\end{center}
\end{figure*}

\begin{figure*}
\begin{center}
\includegraphics[width=0.45\textwidth]{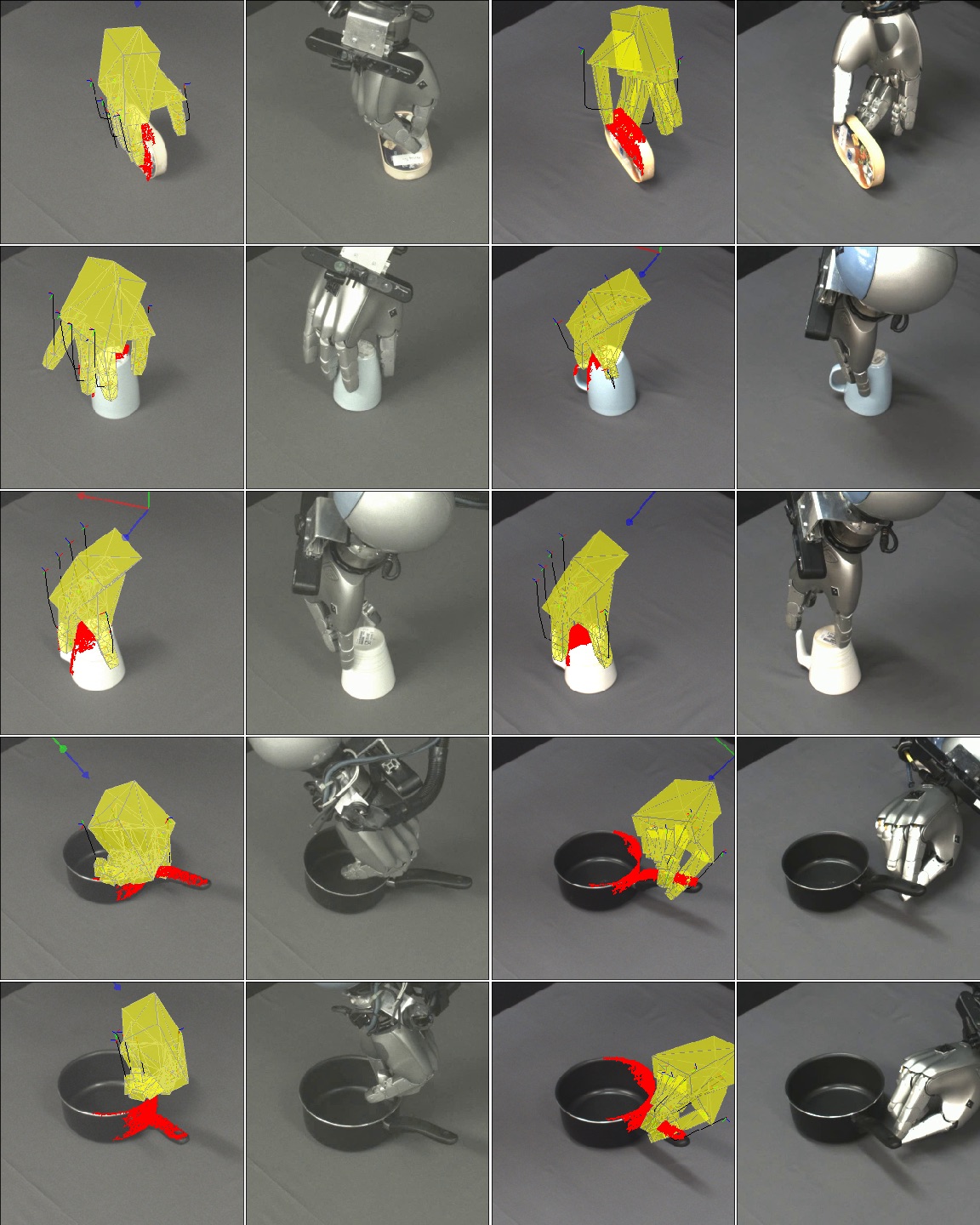}~
\includegraphics[width=0.45\textwidth]{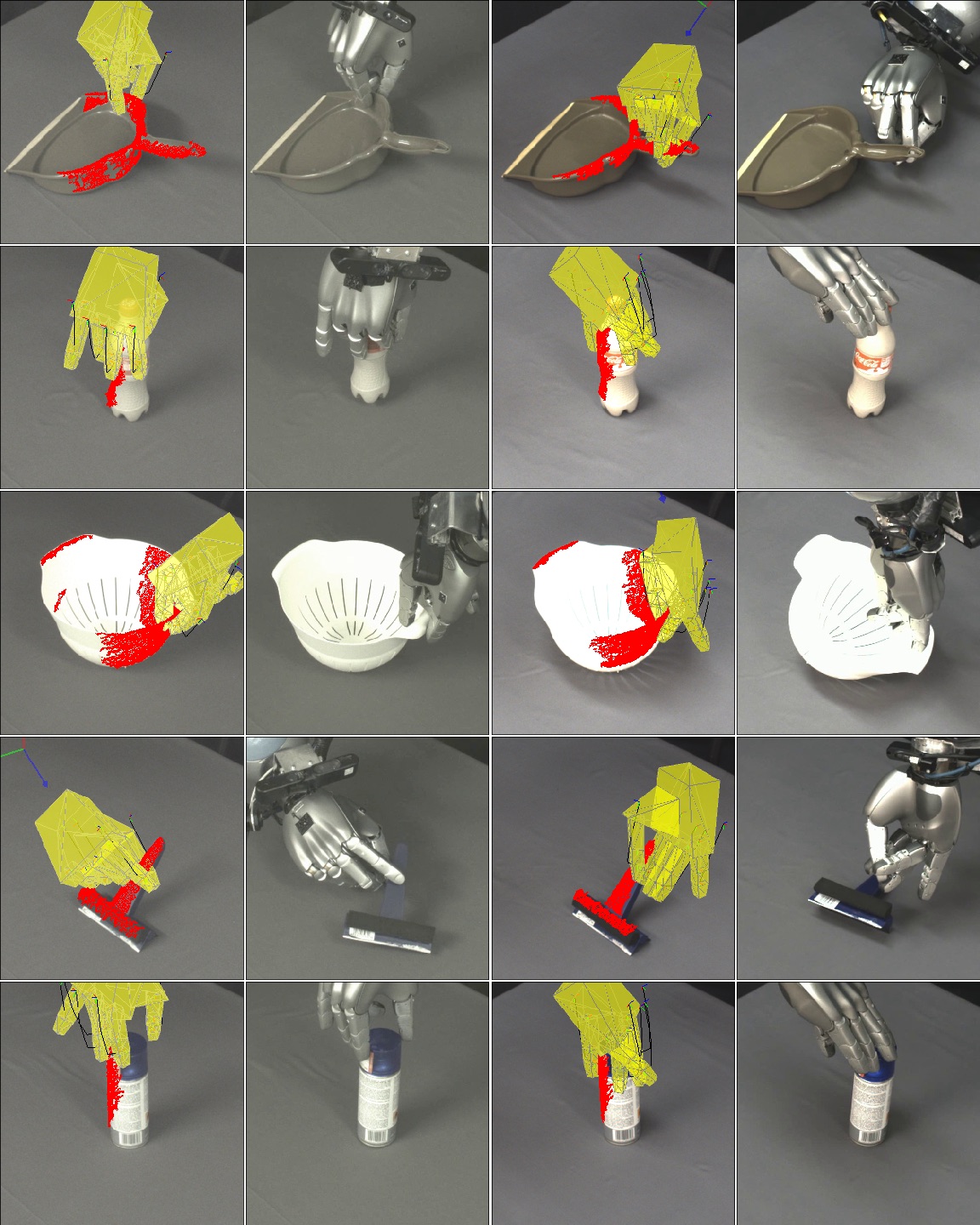}
\caption{V1 vs V4. This shows grasps from methods based on generative model GM1. The V1 grasps are shown in columns 1-2 and 5-6. The corresponding V4 grasps are shown in columns 3-4 and 7-8. The left hand panel shows the cases where both V1 and V4 failed. The right hand panel shows the cases where V1 succeeded and V4 failed. \label{fig:v1fsv4f}}
\end{center}
\end{figure*}


%

\section{Conclusion} 
\label{sec:conclusion}
This paper has presented the first generative-evaluative architecture for dexterous grasping from a single view in which both the generative and evaluative models are learned. Using this architecture the success rate for the top ranked grasp rises from 69.5\% (for V1) to 90.49\% (for V11) on a simulated test set. It also presented a real robot data set where the top ranked grasp success rate rose from 57.1\% (V1) to 87.8\% (V11).

What are the promising lines of enquiry to further improve dexterous grasping of unfamiliar objects? We see three major issues. First, we have assumed no notion of object completion. Humans succeed in grasping in part because we have strong priors on object shape that help complete the missing information. This would enable the deployment of a generative model that exploits a more complete object shape model \cite{kopicki2015ijrr}. Second, our approach is open-loop during execution. For pinch-grasping, deep nets have been shown to learn useful visual servoing policies \cite{morrison18}. However, significant gains will also come from post-grasp force-control strategies, which are largely absent from the literature on grasp learning.  Third, the architectural scheme presented here is essentially that of an actor-critic architecture. This suggests incremental refinement of both the generative model and the evaluative model, perhaps using techniques from reward based learning. We have already shown elsewhere that the GM may be further improved by training from autonomously generated data \cite{kopicki2019}. Data intensive generative models also hold promise \cite{veres2017modeling} and it may be possible to seed them by training with example grasps drawn from a data-efficient model such as that presented here.

\ifCLASSOPTIONcaptionsoff
  \newpage
\fi



%
\bibliographystyle{IEEEtran}
\bibliography{new_total}

%




\begin{IEEEbiography}
[{\includegraphics[width=1in,height=1.25in,clip,keepaspectratio]{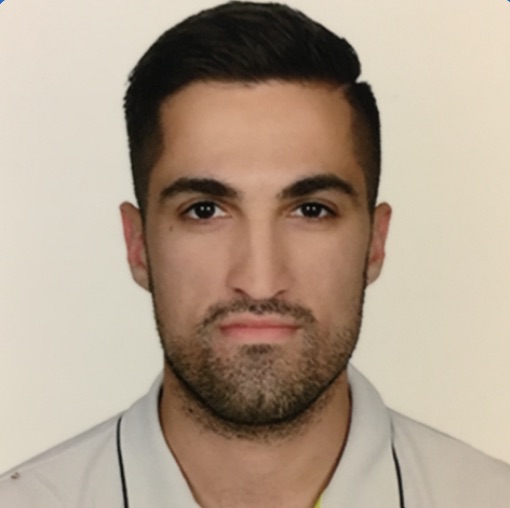}}]{Umit Rusen Aktas}
Umit Rusen Aktas is a Research Engineer at Blue Prism in London. He obtained his PhD in Computer Science from the University of Birmingham in 2018. He received a BSc and MSc in Computer Engineering from METU, Ankara in 2010 and 2013, respectively. He has published two papers at conferences in Computer Vision (ECCV, ICCV). His research interests include machine learning, robot grasping, machine vision and computer graphics. 
\end{IEEEbiography}

\begin{IEEEbiography}
[{\includegraphics[width=1in,height=1.25in,clip,keepaspectratio]{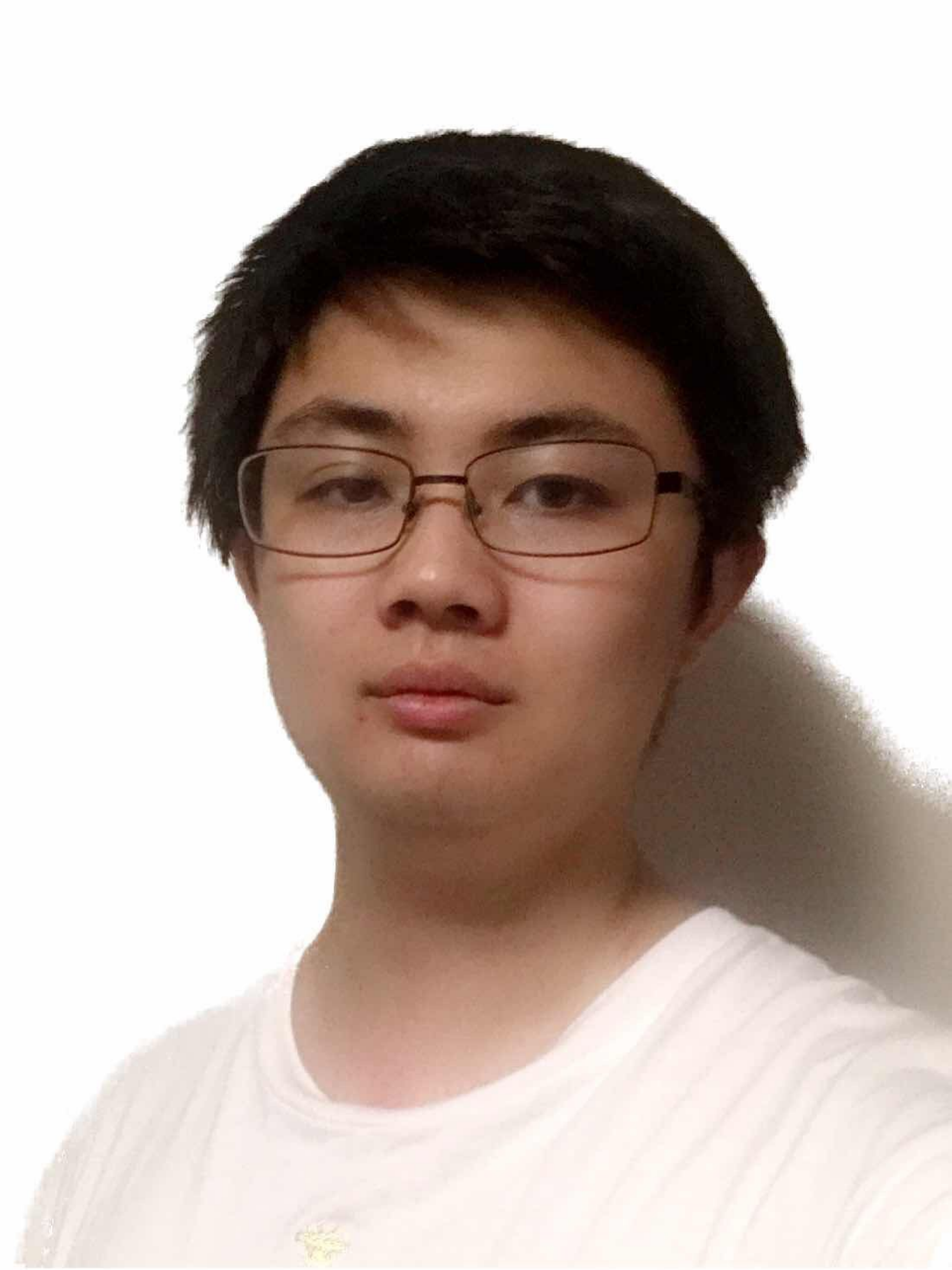}}]{Chao Zhao}
Chao Zhao received the B.S. degree from the University of Electronic Science and Technology of China and an MSc in Computer Science from the University of Birmingham, UK, in 2019. He is currently a Research Engineer at the Beijing Research Institute of China Telecom. His research interests include Artificial Intelligence with both probabilistic and deep learning approaches and its application in robotics.
\end{IEEEbiography}

\begin{IEEEbiography}
[{\includegraphics[width=1in,height=1.25in,clip,keepaspectratio]{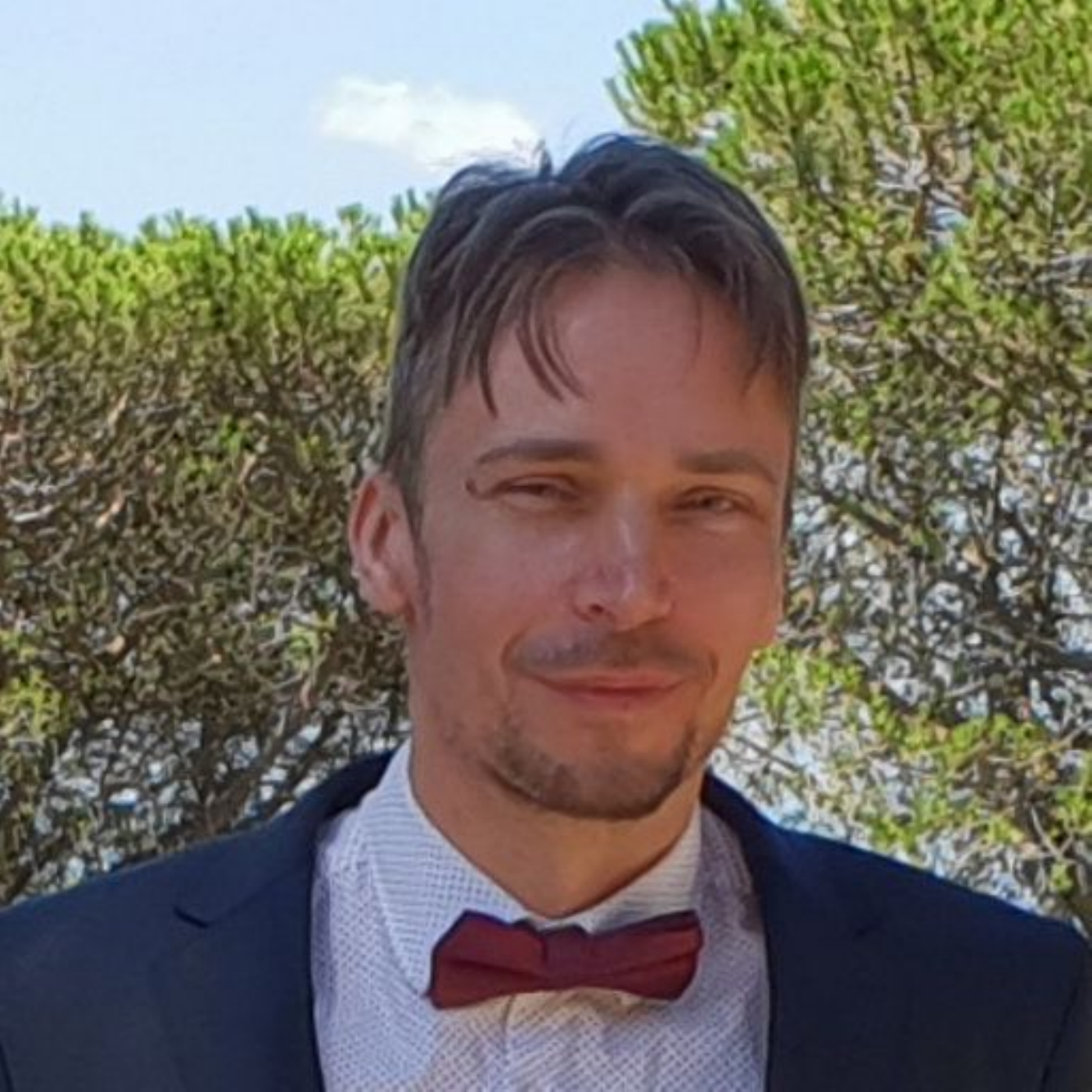}}]{Marek Kopicki}
is a senior research engineer at Dyson. He has a MSc in physics from Adam Mickiewicz University (Poznan), a MSc in Advance Computer Science and Ph.D. in robotics at University of Birmingham in 2010. He has more than twelve years’ experience of developing algorithms and software for robot control and vision. These were used in  FP7 projects CogX, GeRT and PaCMan. He
invented an adaptive grasping algorithm, for which he holds an international patent. He has published more than twenty-five refereed papers.

\end{IEEEbiography}

\begin{IEEEbiography}
[{\includegraphics[width=1in,height=1.25in,clip,keepaspectratio]{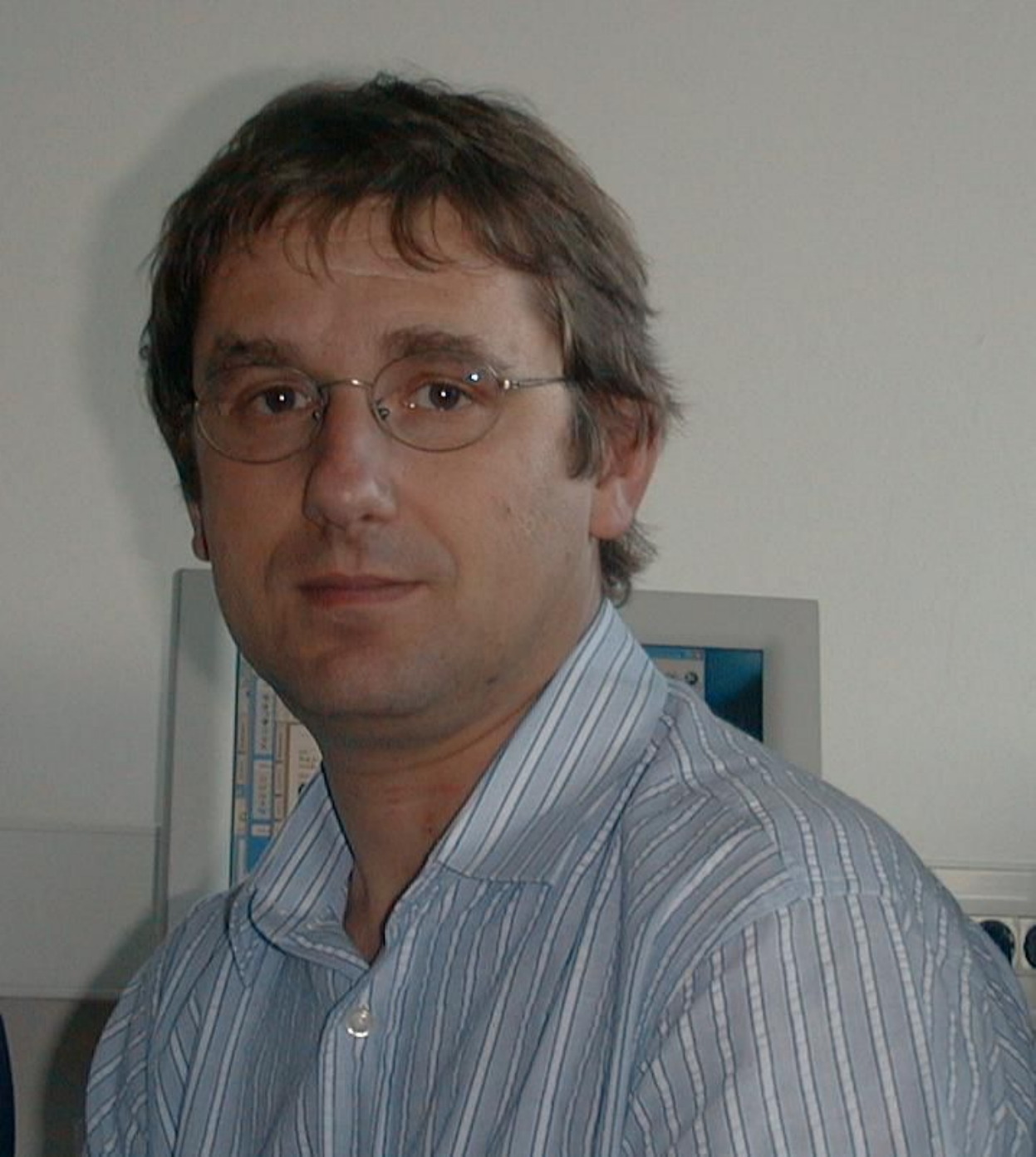}}]{Ales Leonardis}
is a Professor of Robotics at University of Birmingham. He is also an adjunct professor at the Faculty of Computer Science, Graz University of Technology. His research interests include robust and adaptive methods for computer vision, object and scene recognition and categorization, statistical visual learning, 3D object modeling, and biologically motivated vision. He is a fellow of the IAPR and a member of the IEEE and the IEEE Computer Society.
\end{IEEEbiography}

\begin{IEEEbiography}
[{\includegraphics[width=1in,height=1.25in,clip,keepaspectratio]{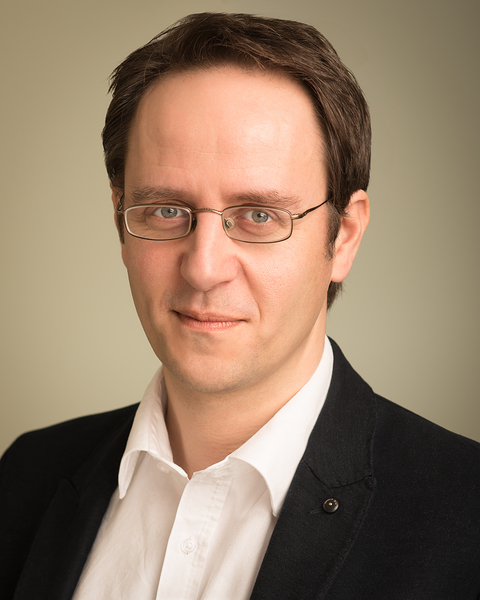}}]{Jeremy L. Wyatt}
received a B.A. in Theology from University of Bristol, an M.Sc. in Artificial Intelligence from University of Sussex, and his Ph.D. in Machine Learning from the University of Edinburgh in 1997. He is Honorary Professor of Robotics and Artificial Intelligence at the University of Birmingham, Birmingham. He has published more than 100 papers, edited three books, received two best paper awards, supervised a BCS distinguished doctoral dissertation, and led a variety of research projects. His research interests include robot task planning, statistical machine learning, and robot manipulation.
\end{IEEEbiography}


\vfill


\end{document}